\documentclass{article}

\PassOptionsToPackage{numbers,sort&compress}{natbib}
\usepackage[eandd, final]{neurips_2026}

\usepackage[utf8]{inputenc} %
\usepackage[T1]{fontenc}    %
\usepackage{url}            %
\usepackage{booktabs}       %
\usepackage{amsfonts}       %
\usepackage{nicefrac}       %
\usepackage{microtype}      %
\usepackage{xcolor}         %

\definecolor{linkcolor}{HTML}{00466B}
\definecolor{citecolor}{HTML}{2F7BA3}
\usepackage[colorlinks=true,linkcolor=linkcolor,citecolor=citecolor,urlcolor=citecolor]{hyperref}
\usepackage{multirow}
\usepackage{longtable}
\usepackage{array}
\usepackage{colortbl}
\usepackage{tabularx}
\usepackage{amssymb}
\usepackage{amsmath}
\usepackage{makecell, adjustbox}
\usepackage{float}
\usepackage{comment}
\usepackage{threeparttable}
\definecolor{myblue}{HTML}{84BBC5}
\definecolor{lightgray}{HTML}{E2E2E2}
\definecolor{sdgray}{gray}{0.4}
\newcommand{\g}{\cellcolor{lightgray}}

\newcommand{\sd}[1]{{\color{sdgray}\,\tiny$\pm$#1}} 

\makeatletter
\renewcommand{\paragraph}{%
  \@startsection{paragraph}{4}%
  {\z@}{0ex \@plus 1ex \@minus .2ex}{-0.3em}%
  {\normalfont\normalsize\bfseries}%
}
\makeatother

\usepackage{tikz}
\usetikzlibrary{positioning, arrows.meta, decorations.pathreplacing, calc}

\title{Where Do We (Not) Need Temporal Context in Low-Resource Video Task Adaptation?}

\author{%
  Luc P.J. Str\"ater \\
  Leiden University \\
  \texttt{l.p.j.strater@liacs.leidenuniv.nl}
  \And
  Hazel Doughty \\
  Leiden University \\
  \texttt{h.r.doughty@liacs.leidenuniv.nl}
}

\begin{document}

\maketitle
\vspace{-0.5em}

\begin{abstract}
\vspace{-0.5em}
  Parameter-efficient finetuning (PEFT) and probing enable adaptation of foundation models using only a small number of trainable parameters. This makes these techniques attractive for video understanding where annotation and computation are expensive. However, video PEFT has focused on adapting image-pretrained models, while standard PEFT methods can also be applied to video representations. These settings are rarely compared and both confine temporal reasoning to a single component of the model, leaving open how temporal context should be distributed across backbone, PEFT, and probe. In this work, we provide a systematic study of model adaptation strategies for video understanding. We evaluate methods across appearance-focused, motion-focused and spatially dense settings, with a particular emphasis on scenarios with limited data where parameter efficiency is most beneficial. Our results provide new insights into PEFT and probing across task types and demonstrate the importance of temporal context allocation for effective video adaptation. Project page: \url{https://lucstrater.com/temporal-context/}.
\end{abstract}

\vspace{-0.5em}
\section{Introduction}
\label{sec:intro}

\vspace{-0.5em}
Foundation models~\citep{bommasani2021opportunities} have become the dominant paradigm in computer vision, offering generalization through internet-scale pretraining~\citep{radford2021learning,simeoni2025dinov3,assran2025v,carreira2024scaling}. Despite learning transferable representations, these models still require adaptation to task-specific objectives, label spaces or domains. This is particularly costly for video, where both annotation and computation are significantly more expensive than for images. As a result, adaptation is increasingly performed through task-specific probing rather than full finetuning~\citep{bardes2024revisiting,chen2025video}. In parallel, parameter-efficient finetuning (PEFT) methods~\citep{houlsby2019parameter,ding2023parameter} adapt models by training only a small number of parameters~\citep{pan2022st}. Yet how these two paradigms interact, and how they should be applied in practice for video understanding, remains unexplored.

Existing work in video PEFT has either adapted image-pretrained models to video~\citep{pan2022st,qing2023disentangling} or evaluated standard PEFT on video-pretrained models~\citep{xin2024v}, without directly comparing the two. Beyond this gap, these works also make different assumptions about where temporal reasoning should occur. Methods built on image-pretrained backbones introduce temporal reasoning only in the adapter~\citep{pan2022st,qing2023disentangling}, while those built on video-pretrained models limit it to the backbone~\citep{xin2024v,yuan2023videoglue,li2024videoeval}. In both cases, the temporal context is limited to one part of the model. Yet there is no reason this design should be optimal. Different tasks may benefit from temporal reasoning at different stages of the adaptation, and the cost of temporal context varies across components. %

In this paper, we present a systematic study of PEFT methods and probing strategies in video understanding. We focus on settings where model adaptation is most valuable: low-resource tasks from specialized domains where data or annotation is inherently limited. Our benchmark spans appearance-focused, motion-focused, and spatially dense tasks with diverse annotation requirements. To study temporal reasoning, we distribute temporal context across the backbone, probe, and PEFT modules, allowing each to operate on a different number of frames, and study the resulting performance-efficiency trade-offs. 
In the low-resource setting, our study reveals four key findings: 
\textbf{(1)} Standard PEFT methods (\textit{e.g.} AdaptFormer~\citep{chen2022adaptformer}) combined with an attentive probe consistently improve over probing alone with little throughput cost. They also outperform or match full finetuning, while training $8\times$ fewer parameters.
\textbf{(2)} Although probe performance differs substantially across tasks, this variability largely disappears once combined with PEFT, allowing a simple attentive probe to replace task-specific decoders.
\textbf{(3)} Temporal modeling is most beneficial for motion tasks, where it is primarily needed in the backbone. For appearance and spatial tasks, the benefit of more frames comes mainly from additional information rather than temporal reasoning.
\textbf{(4)} Beyond the backbone, image-to-video PEFT (\textit{e.g.} ST-Adapter~\citep{pan2022st}) rarely improves over standard PEFT on either image or video backbones, while reducing throughput. Temporal aggregation in the probe can even degrade dense prediction on spatial tasks. Only when a video backbone’s temporal context is reduced can temporal PEFT partially compensate on motion tasks, highlighting a promising direction for efficient video-specific adaptation.

In summary, we (i) provide the first comprehensive evaluation of parameter-efficient adaptation and probing strategies for video, covering both image-pretrained and video-pretrained backbones across diverse tasks and settings. (ii) Introduce temporal context distribution as a new perspective for understanding and improving parameter-efficient model adaptation in video. (iii) Offer concrete adaptation strategies for low-resource video tasks spanning appearance, motion, and spatial understanding.

\vspace{-0.5em}
\section{Related Work}
\label{sec:background}

\vspace{-0.5em}
\noindent\textbf{Video Task Adaptation.} Video understanding commonly relies on large pretrained encoders~\citep{arnab2021vivit,bertasius2021space,tong2022videomae,wang2022internvideo,wang2023videomae,wang2024internvideo2,bardes2024revisiting,carreira2024scaling,assran2025v,chen2025vl,wang2025internvideo}, however adapting them to downstream tasks via end-to-end finetuning is computationally expensive and can overwrite useful priors~\citep{bardes2024revisiting}. This motivates probing strategies, where a lightweight prediction head is trained on a frozen encoder. Probe design varies in how temporal information is modeled. Linear probes aggregate tokens before classification, discarding temporal structure, while attentive probes~\citep{chen2024context,el2024scalable,bardes2024revisiting,psomas2026attention} use cross-attention over spatio-temporal tokens and are a common choice for video~\citep{bardes2024revisiting,assran2025v,simeoni2025dinov3,wang2025internvideo,carreira2024scaling,zoran2025recurrent}. Recent work~\citep{psomas2026attention} improves attentive probe efficiency for classification by removing redundant projections. Other approaches explicitly model time at the probe, enabling adaptation of image encoders to video~\citep{lin2022frozen,qing2023disentangling}. DiST~\citep{qing2023disentangling} adds a lightweight temporal encoder with shallow-to-deep feature fusion emphasizing high-level representations. In contrast, for dense prediction in images, DPT~\citep{ranftl2021vision} fuses features from deep-to-shallow, emphasizing early-layer features. VDA~\citep{chen2025video} extends this with a spatio-temporal head for inter-frame consistency. Despite this diversity, probes are typically developed and evaluated for specific tasks, with limited comparison across tasks or probe types. Moreover, probing and PEFT are studied separately, despite both operating on a frozen backbone. We instead evaluate probe designs jointly with PEFT methods and study how distributing the temporal context between backbone, probe, and PEFT affects adaptation across appearance, motion, and spatially dense video tasks. 

\noindent\textbf{Parameter-Efficient Finetuning.}
While probing adapts pretrained models through the prediction head, PEFT introduces learnable updates within a frozen backbone and is %
complementary to probe-based adaptation~\citep{el2024scalable,psomas2026attention}.
PEFT methods can be divided into four paradigms. 
\textit{Adapters}~\citep{houlsby2019parameter,pfeiffer2021adapterfusion, chen2022adaptformer,luo2023towards,jie2024convolutional,zhao2024dynamic} insert bottleneck modules within transformer blocks, with %
AdaptFormer~\citep{chen2022adaptformer} placing them in parallel with the MLP, rather than after it~\citep{houlsby2019parameter}. \textit{Additive} methods~\citep{lester2021power,li2021prefix,jia2022visual, zeng2024visual, wang2025attention} introduce learnable tokens into the input or intermediate representations, including %
frequency-aware (VFPT~\citep{zeng2024visual})
and low-rank (BPT~\citep{wang2025attention}) variants. \textit{Selective}~\citep{zaken2022bitfit, lian2022scaling,xie2023difffit,basu2024strong} approaches instead update subsets of existing parameters, \textit{e.g.} bias terms (BitFit~\citep{zaken2022bitfit}), normalization (LayerNorm~\citep{basu2024strong}) or per-channel scale and shift (SSF~\citep{lian2022scaling}). %
\textit{Reparameterization} methods~\citep{hu2022lora, zhang2023adalora,jie2023fact,liu2024dora,veeramacheneni2025canonical} inject low-rank residuals into existing weights. For example, LoRA~\citep{hu2022lora} parameterizes the residual as the product of two low-rank matrices, while DoRA~\citep{liu2024dora} further decomposes pretrained weights into magnitude and direction components. %
While these methods are well-studied in image and language, their behavior for video remains less clear.

Existing video-specific PEFT methods~\citep{lin2022frozen,pan2022st,park2023dual,yangaim, ponbagavathi2026frame2freq, liu2026static} instead focus on image-to-video transfer, where a frozen image backbone~\citep{radford2021learning,caron2021emerging,oquab2023dinov2,simeoni2025dinov3,tschannen2025siglip} is augmented with temporal modules. %
For example, ST-Adapter~\citep{pan2022st} inserts a temporal 3D convolution into the adapter bottleneck. %
These approaches are typically trained on large-scale video datasets and adopt a fixed temporal configuration, where temporal reasoning is introduced only through the adaptation modules while the backbone remains static.
As a result, they are tailored to large-scale action recognition. Crucially, it remains unclear how these image-to-video approaches compare to applying standard PEFT methods to video backbones. In contrast, we evaluate PEFT methods across diverse video tasks in the low-data regime and study how temporal context should be distributed across the backbone and adaptation modules. %

\noindent\textbf{Adaptation Benchmarks.}
Model adaptation is particularly useful in low-resource settings, where limited data makes full finetuning undesirable. This is recognized in images, where models pretrained on large-scale image datasets~\citep{deng2009imagenet, ridnik2021imagenet,schuhmann2022laion} are evaluated on low-resource benchmarks~\citep{zhai2019large,kornblith2019better}. These settings typically involve limited data, fine-grained distinctions, and domain shift between the pretraining and downstream tasks~\citep{zhang2024low}. In contrast, video adaptation is typically evaluated in the opposite regime: large-scale datasets, coarse-grained action recognition and web-domain videos closely aligned with the pretraining data, providing limited insight into generalization beyond these settings~\citep{thoker2022severe}. While recent works begin to consider low-resource and specialized settings~\citep{thoker2025severe++, li2024videoeval, hasson2025scivid}, they focus on backbone design and pretraining, rather than parameter-efficient methods. Similarly, streaming benchmarks for video LLMs~\citep{Niu_2025_CVPR,lin2026streamingbench} lack training data, but evaluate models zero-shot rather than through adaptation. Systematic evaluations of PEFT do exist in other domains~\citep{ding2023parameter,xin2024parameter,mai2025lessons, belanec2026peft}, but these findings do not directly transfer to video, 
where temporal structure introduces an additional dimension for adaptation.

Closest to our study is V-PETL~\citep{xin2024v}, which extends its analysis of PEFT on images to three video datasets (K400~\citep{kay2017kinetics}, SSv2~\citep{goyal2017something}, HMDB~\citep{kuehne2011hmdb}). However, these evaluations focus on action recognition benchmarks that are either large-scale or closely aligned with pretraining data, limiting insight into low-resource regimes and the role of temporal context.  %
In contrast, we benchmark adaptation methods on naturally low-resource video tasks spanning appearance, motion and spatial understanding and study how temporal context affects adaptation across the backbone, probe and PEFT modules.

\vspace{-0.5em}
\section{Distributing Temporal Context}
\label{sec:method}
\vspace{-0.6em}

\begin{figure}[t]
    \centering
    \includegraphics[width=0.9\linewidth]{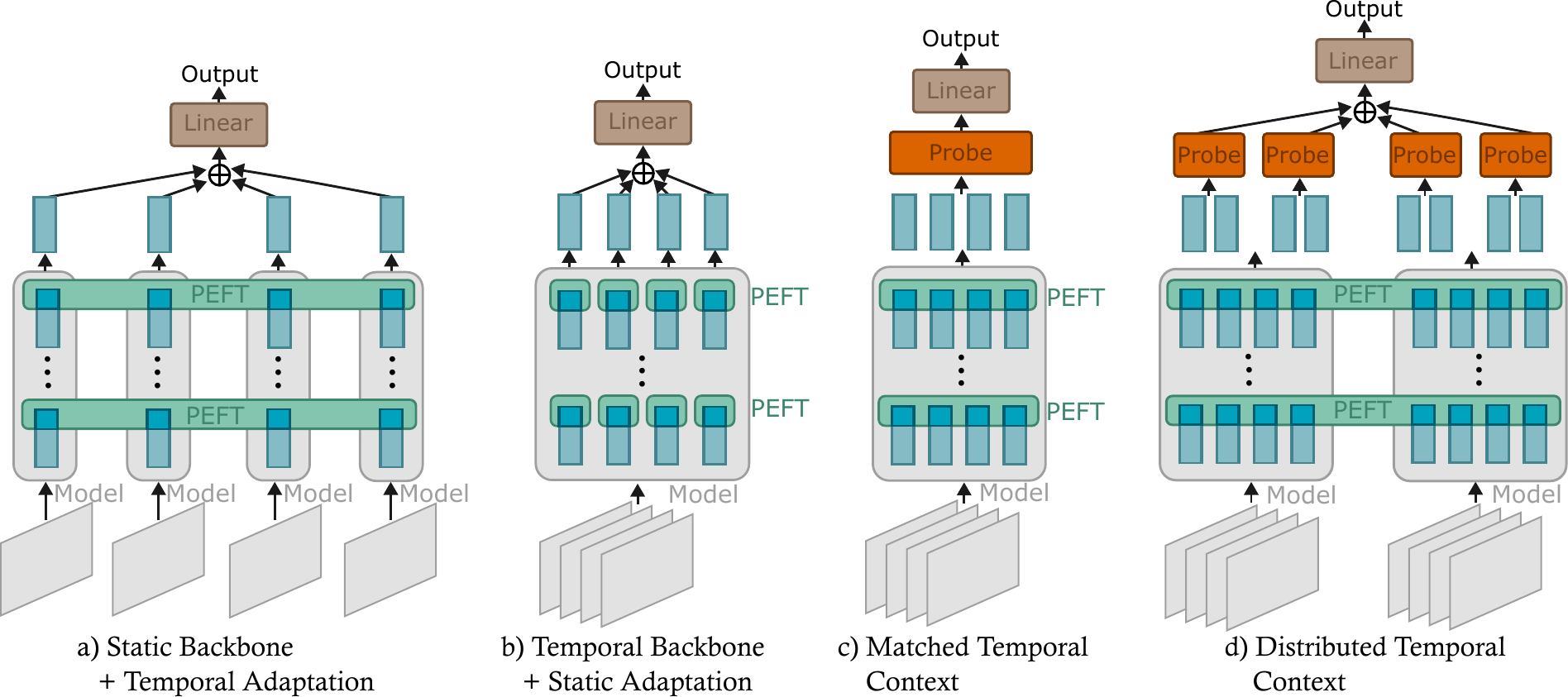}
    \vspace{-0.5em}
    \caption{\textbf{Distributing Temporal Context.} Existing works place temporal modeling in a single component: either in the PEFT modules (a), where a frame-wise backbone is combined with temporal adaptation, or in the backbone (b), where temporally aware representations are adapted by static PEFT and a linear classifier. We instead allow temporal modeling across backbone, PEFT, and probe simultaneously (c) and study how temporal context should be distributed across components (d).}
    \vspace{-1em}
    \label{fig:method-decoupling}
\end{figure}

Video adaptation requires reasoning over multiple frames, making a central design choice not just \textit{whether} to model temporal information, but \textit{where} this temporal reasoning should occur. Existing approaches make this choice implicitly by assigning temporal modeling to a single component. In image-to-video adaptation~\citep{pan2022st}, the backbone operates on individual frames and temporal reasoning is only introduced through the PEFT (Figure~\ref{fig:method-decoupling}a). In contrast, standard PEFT or probing methods can be applied directly to video backbones~\citep{xin2024v} (Figure~\ref{fig:method-decoupling}b). 
Despite their differences, both concentrate temporal capacity only in a single component. This leaves a broader design space underexplored in which multiple components may benefit from temporal reasoning at different temporal scales, raising the question of
\textit{how temporal context should be distributed across the backbone, PEFT modules, and probe for maximum efficiency and performance?}

In this study, we consider a setting where temporal modeling can be applied across the backbone, PEFT and probe (Figure~\ref{fig:method-decoupling}c). This treats temporal context as a flexible resource that can be distributed across components with different temporal extents (Figure~\ref{fig:method-decoupling}d), enabling a systematic study of where performance gains arise. We begin with the matched setting (Section~\ref{sec:matched}), where all components operate on the same temporal context and progressively relax this constraint by distributing the temporal context differently across probe (Section~\ref{sec:method_decoupled_probe}) and PEFT modules (Section~\ref{sec:method_decoupled_peft}).

\vspace{-0.5em}
\subsection{Matched Temporal Context}
\vspace{-0.3em}
\label{sec:matched}
We first consider the matched setting, where the backbone, probe and PEFT modules all operate on the same temporal context.
Given a video $X \in \mathbb{R}^{T \times H \times W \times 3}$ where $T$ is the number of frames,  a video encoder $f_\theta$ maps the clip into spatio-temporal tokens and produces representations $Z$:
\vspace{-0.3em}
\begin{equation}
Z = f_\theta(X) \in \mathbb{R}^{T \cdot N_s \times D}
\end{equation}
where $N_s$ is the number of spatial tokens per frame and $D$ is the embedding dimension, giving a total of $N{=}T{\cdot}N_s$ tokens. We use a ViT~\citep{dosovitskiy2020image}, thus $f_\theta$ consists of a stack of $L$ transformer blocks. At layer $\ell$, given intermediate features $h^{\ell} \in \mathbb{R}^{N \times D}$, the transformer applies multi-head self-attention (MSA) followed by an MLP. For adaptation, both modules are augmented with PEFT operators $\mathcal{P}_{\mathrm{MSA}}^{\ell}$ and $\mathcal{P}_{\mathrm{MLP}}^{\ell}$, which introduce a small set of trainable parameters:
\begin{align}
\tilde{h}^{\ell} &= h^{\ell} + \mathcal{P}_{\mathrm{MSA}}^{\ell}\!\left(h^{\ell},\, \mathrm{MSA}(h^{\ell})\right) \qquad
h^{\ell+1} = \tilde{h}^{\ell} + \mathcal{P}_{\mathrm{MLP}}^{\ell}\!\left(\tilde{h}^{\ell},\, \mathrm{MLP}(\tilde{h}^{\ell})\right) \label{eq3}
\end{align}
To obtain a final prediction, a probe $g_\phi$ aggregates the $N$ output tokens in the representation $Z{=}h^{L}$ into a task prediction $\hat{y}$. In this setting the probe, PEFT modules, and backbone all operate on the same temporal context, 
receiving all tokens from the $T$ input frames. We denote this matched context as:
\begin{equation}
T_{\mathrm{backbone}} = T_{\mathrm{probe}} = T_{\mathrm{peft}} =  T.
\end{equation}

\vspace{-0.3em}
\subsection{Distributing Probe Temporal Context}
\vspace{-0.5em}
\label{sec:method_decoupled_probe}
In the matched setting, the probe operates on all tokens and shares the backbone's temporal context. As increasing temporal context in the backbone is computationally expensive, we consider instead varying it at the probe, either increasing or reducing it. %

\textbf{Longer Probe Context Than Backbone.}
The probe can operate on a longer temporal context by aggregating tokens from multiple backbone instances, as in VJEPA \citep{bardes2024revisiting}. This avoids the quadratic computation for full spatio-temporal attention in the backbone. Given an input clip of $T_{\mathrm{probe}}$ frames and a backbone using $T_{\mathrm{backbone}} < T_{\mathrm{probe}}$, we divide the video into $K{=}T_{\mathrm{probe}} / T_{\mathrm{backbone}}$ segments $X_1, \ldots, X_K$, apply the \emph{same} backbone to each and concatenate the output:
\begin{equation}
Z_k = f_\theta(X_k) \in \mathbb{R}^{T_{\mathrm{backbone}} \cdot N_s \times D}, \quad Z = [Z_1; Z_2; \ldots; Z_K] \in \mathbb{R}^{T_{\mathrm{probe}} \cdot N_s \times D}, \quad k = 1, \ldots, K.
\end{equation}
The probe then attends over $Z$ to give the task prediction $\hat{y}{=}g_\phi(Z)$.

\textbf{Shorter Probe Context Than Backbone.}
Conversely, the probe can have a shorter temporal context, $T_{\mathrm{probe}}{<}T_{\mathrm{backbone}}$,  %
when the backbone already captures sufficient temporal information. %
Given backbone output tokens $Z{\in}\mathbb{R}^{T_{\mathrm{backbone}} \cdot N_s \times D}$, we partition the temporal axis into $K$ segments
$Z{=}[Z^{1}; Z^{2}; \ldots; Z^{K}]$ where $Z^{k}{\in} \mathbb{R}^{T_{\mathrm{probe}} \cdot N_s \times D}$.
The \textit{same} probe $g_\phi$ is applied independently to each segment, producing per-segment outputs $\hat{y}^{k}{=}g_\phi(Z^{k})$. For dense tasks, these are concatenated temporally; for classification, they are average-pooled before a linear prediction head.

\vspace{-0.3em}
\subsection{Distributing PEFT Temporal Context}
\vspace{-0.5em}
\label{sec:method_decoupled_peft}

The same distribution can be applied to PEFT modules. While Section~\ref{sec:method_decoupled_probe} varies the temporal context at the probe, here we vary it within the model. In the matched setting, $\mathcal{P}_{\mathrm{MSA}}^{\ell}$ and $\mathcal{P}_{\mathrm{MLP}}^{\ell}$ inherit the backbone's temporal context $T_{\mathrm{backbone}}$. We instead allow the PEFT operators to act over their own temporal context $T_{\mathrm{peft}}$, considering both longer and shorter cases.

\textbf{Longer PEFT Context Than Backbone.}
This regime %
allows PEFT modules to mix information across frames that are never processed jointly by the backbone, enabling long-range temporal interaction at adapter cost. Given a clip of $T_{\mathrm{peft}}$ frames and a backbone operating on $T_{\mathrm{backbone}}$, we process $K{=}T_{\mathrm{peft}} / T_{\mathrm{backbone}}$ segments to obtain per-layer intermediate features $h_k^{\ell} \in \mathbb{R}^{T_{\mathrm{backbone}} \cdot N_s \times D}$ for $k{=}1, \ldots, K$. We concatenate these along the temporal axis
$H^{\ell} = [h_1^{\ell}; \ldots; h_K^{\ell}] \in \mathbb{R}^{T_{\mathrm{peft}} \cdot N_s \times D}$.
The PEFT operator acts on $H^{\ell}$, enabling temporal interaction across segments, while the MSA continues to operate per segment. The result is split back into segments (denoted by $[\cdot]$) to give:
\begin{equation}
\tilde{h}_k^{\ell} = h_k^{\ell} + \left[\mathcal{P}_{\mathrm{MSA}}^{\ell}\!\left(H^{\ell},\, \mathrm{MSA}(h_k^{\ell})\right)\right]_k.
\end{equation}
The $\mathcal{P}_{\mathrm{MLP}}^{\ell}$ case is analogous, %
using the concatenated features $\tilde{H}^{\ell} = [\tilde{h}_1^{\ell}; \ldots; \tilde{h}_K^{\ell}]$. %

\textbf{Shorter PEFT Context Than Backbone.}
Conversely, PEFT can operate on a shorter temporal context $T_{\mathrm{peft}}{<}T_{\mathrm{backbone}}$.
This restricts temporal mixing within the adapter to local temporal patterns while longer-range dependencies are handled by the backbone. We partition the backbone features along the temporal axis into $K{=} T_{\mathrm{backbone}} / T_{\mathrm{peft}}$ segments:
$h^{\ell}{=}[h^{\ell,1}; \ldots; h^{\ell,K}]$ where $h^{\ell,k} \in \mathbb{R}^{T_{\mathrm{peft}} \cdot N_s \times D}$.
The \textit{same} PEFT operator is applied independently to each segment, while the MSA acts at the full backbone context. The resulting updates are concatenated temporally to give:
\begin{equation}
\tilde{h}^{\ell} = h^{\ell} + \big[\mathcal{P}_{\mathrm{MSA}}^{\ell}\!\left(h^{\ell,1},\, [\mathrm{MSA}(h^{\ell})]_1\right);\, \ldots;\, \mathcal{P}_{\mathrm{MSA}}^{\ell}\!\left(h^{\ell,K},\, [\mathrm{MSA}(h^{\ell})]_K\right)\big].
\end{equation}
The $\mathcal{P}_{\mathrm{MLP}}^{\ell}$ case is analogous, partitioning both $\tilde{h}^{\ell}$ and $\mathrm{MLP}(\tilde{h}^{\ell})$ into segments.

\vspace{-0.5em}
\section{Low-Resource Video Evaluation}
\label{sec:benchmark}
\vspace{-0.5em}
To study distributed temporal context, we require a benchmark that exposes when different adaptation and temporal modeling strategies matter.
Existing video adaptation~\citep{xin2024v,yuan2023videoglue,li2024videoeval} focuses on large-scale recognition where abundant data reduces the need for parameter-efficient methods. PEFT is most useful in low-resource settings~\citep{mai2025lessons}, where specialized domains make adaptation necessary and limited data makes full finetuning prone to overfitting. We therefore define criteria for intrinsically low-resource video tasks (Section~\ref{sec:selection_criteria}) and use them to construct our ScarceVid benchmark (Section~\ref{sec:tasks_datasets}).

\vspace{-0.2em}
\subsection{Selection Criteria}
\label{sec:selection_criteria}
\vspace{-0.3em}
We cannot simply subsample large-scale datasets, as this fails to reproduce the challenges of genuinely scarce data~\citep{zhang2024low}. We instead select datasets that are intrinsically limited in size with four criteria. %
First, we prioritize settings where \textit{data scarcity} is structural such as healthcare~\citep{nurvid}, wildlife monitoring~\citep{mammalps}, and per-pixel annotation~\citep{vspw}. Second, we ensure \emph{capability coverage} across appearance, motion, and spatial video understanding. %
Third, we span a range of \emph{annotation densities}, from video-level labels to per-pixel annotations, as output granularity may interact with temporal context. 
 Finally, we include diversity in \emph{viewpoint} and \emph{domain} to avoid conclusions tied to a single visual regime.

\vspace{-0.2em}
\subsection{Tasks and Datasets}
\vspace{-0.3em}
\label{sec:tasks_datasets}

Our ScarceVid benchmark spans six datasets covering appearance, motion, and spatial understanding. Table~\ref{tab:benchmark} summarizes key statistics; here we briefly describe the tasks.

\begin{table}
\centering

\caption{\textbf{The ScarceVid Benchmark} covers low-resource appearance, motion, and spatial video understanding tasks across diverse annotation granularities, viewpoints, and domains.}
\vspace{-0.5em}
\label{tab:benchmark}
\begin{threeparttable}
\renewcommand{\arraystretch}{1.15}
\setlength{\tabcolsep}{5pt}
\scriptsize
\begin{tabular}{
  @{}l
  cc
  cc
  cc@{}
}
\toprule
& \multicolumn{2}{c}{\cellcolor{myblue}\textbf{Appearance}}
& \multicolumn{2}{c}{\cellcolor{myblue}\textbf{Motion}}
& \multicolumn{2}{c}{\cellcolor{myblue}\textbf{Spatial}} \\
\cmidrule(lr){2-3}\cmidrule(lr){4-5}\cmidrule(lr){6-7}
& \makecell[c]{\textbf{CAER}}
& \makecell[c]{\textbf{NurViD}}
& \makecell[c]{\textbf{IndustReal}}
& \makecell[c]{\textbf{MammAlps}}
& \makecell[c]{\textbf{ScanNet}}
& \makecell[c]{\textbf{VSPW}} \\
\midrule
\rowcolor{lightgray}
\textbf{Task}
 & \parbox[c]{1.4cm}{\centering Emotion\\Analysis}
 & \parbox[c]{1.4cm}{\centering Procedure\\Classification}
 & \parbox[c]{1.4cm}{\centering Step\\Recognition}
 & \parbox[c]{1.4cm}{\centering Behavior\\Recognition}
 & \parbox[c]{1.4cm}{\centering Depth\\Estimation}
 & \parbox[c]{1.4cm}{\centering Semantic\\Segmentation} \\
\textbf{Viewpoint}
 & 3\textsuperscript{rd}
 & 3\textsuperscript{rd}
 & 1\textsuperscript{st}
 & 3\textsuperscript{rd}
 & 1\textsuperscript{st}
 & 3\textsuperscript{rd} \\
\rowcolor{lightgray}
\textbf{Domain}
 & Television
 & Medical
 & Assembly
 & Wildlife
 & Indoor
 & Web \\
\textbf{Classes}
 & 7 & 51 & 75 & 11 & -- & 124 \\
\rowcolor{lightgray}
\textbf{Train}
 & 9{,}222 & 3{,}899 & 3{,}667 & 4{,}205 & 1{,}201 & 2{,}806 \\
\textbf{Validation}
 & 1{,}316 & 586 & 1{,}928 & 686 & 312 & 343 \\
\rowcolor{lightgray}
\textbf{Test}
 & 2{,}637 & 1{,}121 & 3{,}678 & 1{,}244 & 100 & 387\textsuperscript{\dag} \\
\bottomrule
\end{tabular}
\begin{tablenotes}
\footnotesize
\item[\textsuperscript{\dag}VSPW test labels are not public; we evaluate on the validation set.]
\end{tablenotes}
\end{threeparttable}

\vspace{-1.2em}
\end{table}

\paragraph{Appearance.}
CAER~\citep{caer} contains short TV clips, labeled with one of seven emotions. %
Recognition is driven by facial expression rather than motion. %
NurViD~\citep{nurvid} targets nursing procedure classification. The 51 procedures are visually similar, involving a clinician interacting with a patient or equipment, making this a fine-grained appearance task. %

\paragraph{Motion.}
IndustReal~\citep{industreal} contains egocentric videos of industrial assembly, where each clip is labeled with one of 75 steps. Distinguishing steps requires modeling how the scene evolves over time. 
MammAlps~\citep{mammalps} contains multi-camera wildlife footage with 11 behaviors (e.g. chasing, foraging). These unfold over seconds and are visually subtle, making this a temporally demanding task. %

\paragraph{Spatial.}
ScanNet~\citep{scannet} consists of indoor RGB-D scans, from which we use RGB frames for per-pixel depth regression. Data collection requires specialized hardware, limiting scale. 
VSPW~\citep{vspw} provides per-frame semantic segmentation labels across 124 categories on third-person web video. 
Annotations are hand-drawn on each frame, making them labor-intensive. Together these tasks require dense spatial predictions, allowing us to study how temporal context interacts with per-pixel prediction.

\vspace{-0.5em}
\section{Empirical Study of Video Model Adaptation}
\label{sec:experiments}
\vspace{-0.5em}
We study model adaptation across the tasks in our low-resource video benchmark. Our experiments analyze when to adapt an image or video model (Section~\ref{sec:imagevsvideo}), which PEFT  (Section~\ref{sec:exp_peft}) and probe methods (Section~\ref{sec:exp_probe}) work best and how temporal context should be distributed across input, model, probe and PEFT (Sections~\ref{sec:exp_uniform_input} and~\ref{sec:exp_context}). Full experiment details are provided in Appendix~\ref{app:experiment_details}.

\vspace{-0.3em}
\subsection{When to Adapt an Image Model vs. a Video Model? } \label{sec:imagevsvideo}
\vspace{-0.3em}
\textbf{Experimental Setup.} We consider four pretraining paradigms. For image-only self-supervision, we use DINOv3~\citep{simeoni2025dinov3}, which learns via unsupervised self-distillation. SigLIP2~\citep{tschannen2025siglip} represents image-text pretraining, aligning images and captions with a sigmoid contrastive objective. VJEPA2~\citep{assran2025v} represents video self-supervision, performing masked video modeling in the latent feature space. InternVideo-Next~\citep{wang2025internvideo} combines image-text and video self-supervision by aligning intermediate representations to a frozen SigLIP2 teacher before performing masked video prediction. Pretraining scales differ substantially with SigLIP2 trained on 10B images, DINOv3 on 1.7B images, VJEPA2 on 22M videos and images, and InternVideo-Next on 1.1M videos. For each backbone, we evaluate both an attentive probe alone and in combination with several representative PEFT approaches: ST-Adapter (image-to-video)~\citep{pan2022st}, AdaptFormer~\citep{chen2022adaptformer}, and LoRA~\citep{hu2022lora}. We use the ViT-L version of each backbone as it is the only variant available across all models. Results are shown in Table~\ref{tab:backbones}.

\begin{table}[t]
\centering
\caption{
\textbf{Adapting Image vs.\ Video Models.} A video backbone performs best on appearance and motion tasks despite orders of magnitude fewer pretraining samples; DINOv3 leads on spatial tasks. PEFT reduces the differences between backbones. \textbf{Best} and \underline{best per backbone} are highlighted.}
\vspace{-0.5em}
\label{tab:backbones}
\renewcommand{\arraystretch}{1.15}
\setlength{\tabcolsep}{5pt}
\begin{adjustbox}{max width=\textwidth}
\scriptsize
\begin{tabular}{
  @{}ll
  cc
  cc
  cc@{}
}
\toprule
& & \multicolumn{2}{c}{\cellcolor{myblue}\textbf{Appearance}}
& \multicolumn{2}{c}{\cellcolor{myblue}\textbf{Motion}}
& \multicolumn{2}{c}{\cellcolor{myblue}\textbf{Spatial}}
 \\
\cmidrule(lr){3-4}\cmidrule(lr){5-6}\cmidrule(lr){7-8}
& & \makecell[c]{\textbf{CAER}\\\textit{\scriptsize top-1 $\uparrow$}}
& \makecell[c]{\textbf{NurViD}\\\textit{\scriptsize top-1 $\uparrow$}}
& \makecell[c]{\textbf{IndustReal}\\\textit{\scriptsize top-1 $\uparrow$}}
& \makecell[c]{\textbf{MammAlps}\\\textit{\scriptsize top-1 $\uparrow$}}
& \makecell[c]{\textbf{ScanNet}\\\textit{\scriptsize AbsRel $\downarrow$}}
& \makecell[c]{\textbf{VSPW}\\\textit{\scriptsize mIoU $\uparrow$}}\\
\midrule
\multirow{4}{*}{\rotatebox[origin=c]{90}{\textbf{SigLIP2}}}
 & \g Attentive Probe & \g 52.1 & \g 79.0 & \g 47.4 & \g 66.9 & \g 0.237 & \g 33.8  \\
 & LoRA     & \underline{63.0} & 87.0 & \underline{67.7} & \underline{73.4} & 0.136 & 53.5  \\
 & \g AdaptFormer          &\g 61.2 &\g 87.1 &\g 64.2 &\g 72.4 &\g \underline{0.122} &\g 54.9  \\
  &  ST-Adapter           & 62.5 & \underline{87.5} & 66.9 & 72.4 & 0.131 & \underline{55.7}  \\

\midrule
\multirow{4}{*}{\rotatebox[origin=c]{90}{\textbf{DINOv3}}}
 & \g Attentive Probe & \g 58.9 & \g 85.2 &\g  58.1 &\g  71.5 &\g  0.120 &\g  56.4 \\
 & LoRA               & \underline{62.8} & \underline{87.3} & \underline{69.8} & 72.6 & 0.111 & 60.6 \\
 &\g  AdaptFormer    & \g 61.8 & \g 87.1 & \g 67.5 & \g 72.6 & \g \underline{\textbf{0.092}} & \g \underline{\textbf{61.2}} \\
  &  ST-Adapter         & 62.4 & 86.9 & 69.2 & \underline{75.6} & 0.106 & 61.1  \\
\midrule
\multirow{3}{*}{\rotatebox[origin=c]{90}{\textbf{VJEPA2}}}
 & \g Attentive Probe & \g 54.3 & \g 82.4 & \g 60.2 & \g 74.0 & \g 0.158 & \g 35.4 \\
 & LoRA               & \underline{59.1} & \underline{84.0} & \underline{67.6} & 77.1 & 0.131 & \underline{46.9} \\
 & \g AdaptFormer    & \g 57.8 & \g 83.8 & \g 64.4 & \g \underline{\textbf{77.2}} & \g \underline{0.128} & \g 45.7 \\
 \midrule
\multirow{3}{*}{\rotatebox[origin=c]{90}{\textbf{IV-Next}}}
 & \g Attentive Probe & \g 62.7 & \g 88.5 & \g 67.0 & \g 74.0 & \g 0.135 & \g 47.3   \\
 & LoRA               & 63.1 & 88.5 & \underline{\textbf{72.8}} & 75.9 & 0.142 & 55.2  \\
 & \g AdaptFormer    & \g \underline{\textbf{63.4}} & \g \underline{\textbf{88.6}} & \g 70.0 & \g \underline{76.3} & \g \underline{0.118} & \g \underline{57.6}   \\
\bottomrule
\end{tabular}
\end{adjustbox}
\vspace{-0.5em}
\end{table}

\textbf{Results.} The differences between backbones are large with attentive probing alone, but are substantially reduced when combined with PEFT. In both settings, a video backbone performs best on every appearance and motion dataset, despite being pretrained on orders of magnitude fewer samples than the image backbones. On spatial tasks, DINOv3 performs best, while InternVideo-Next outperforms the image-text model SigLIP2. DINOv3's lead over video backbones is likely due to its larger pretraining size and its Gram anchoring loss. This loss preserves the local spatial structure of patch features. Neither the video backbones nor SigLIP2 uses such a loss.

In terms of adaptation methods, adding temporal modeling to the adapter (e.g. ST-Adapter) does not consistently improve over standard PEFT (AdaptFormer, LoRA) on image backbones. This suggests that with limited adaptation data, temporal reasoning is better modeled in the backbone than in the learned adapter. Additionally, standard PEFT applied to a video-pretrained backbone outperforms image-to-video adaptation on most datasets, suggesting that future research should focus on video-specific adaptation of video backbones rather than image-to-video transfer. Based on these results, we use InternVideo-Next in all subsequent experiments.

\textbf{Conclusion.} A video backbone is preferable for appearance and motion tasks, DINOv3 for spatial tasks. Combining attentive probing with PEFT reduces the differences between backbones. However, adding temporal modeling to the PEFT method on image backbones is not beneficial on most tasks.

\vspace{-0.3em}
\subsection{Which PEFT Methods Work Best for Video?} \label{sec:video_model_peft}
\vspace{-0.4em}
\label{sec:exp_peft}
\textbf{Experimental Setup.} We compare ten PEFT methods spanning four paradigms:
selective (BitFit~\citep{zaken2022bitfit}, LayerNorm~\citep{basu2024strong}, SSF~\citep{lian2022scaling}), additive (VPT~\citep{jia2022visual}, VFPT~\citep{zeng2024visual}, BPT~\citep{wang2025attention}), adapter (ST-Adapter~\citep{pan2022st}, AdaptFormer~\citep{chen2022adaptformer}), and low-rank (LoRA~\citep{hu2022lora}, DoRA~\citep{liu2024dora}). Each is combined with an attentive probe and compared to full finetuning and attentive probing alone in Table~\ref{tab:peft_results}. We use InternVideo-Next ViT-B as the backbone. ViT-B vs.\ ViT-L comparisons are provided in Appendix~\ref{app:backbone_size} and peak inference memory in Appendix~\ref{app:inferece_mem}.

\textbf{Results.} Most PEFT methods improve over the attentive probe alone, with most methods also exceeding full finetuning on spatial tasks. AdaptFormer is the most robust choice across tasks, as it consistently outperforms attentive probing and matches or exceeds full finetuning. This performance comes with 8$\times$ fewer parameters than full finetuning and a 3\% reduction in throughput relative to the attentive probe. Beyond this, the differences between PEFT methods are often small once seed variance is taken into account. On appearance tasks, most methods perform similarly. On motion tasks, LoRA performs best on IndustReal, but the high variance on MammAlps prevents a clear ranking. On spatial tasks, however, the lower variance reveals a consistent advantage of adapter methods over low-rank methods.

Adding temporal modeling to the adapter (ST-Adapter) provides no clear gains over AdaptFormer given the variation across seeds, while requiring additional trainable parameters and reducing throughput. Additive methods are outperformed by adapter methods on most tasks, suggesting that prompting is a weaker adaptation mechanism for low-resource video tasks. In terms of training efficiency, selective methods require the fewest additional parameters, with SSF performing best among them and even achieving the best result overall on ScanNet. In terms of inference efficiency, SSF and the low-rank methods can be combined with the backbone weights at inference time and thus match attentive probing. More broadly, conclusions from other domains do not necessarily transfer to video; for example, DoRA does not outperform LoRA and VFPT does not outperform VPT, despite being more recent.

\textbf{Conclusion.} AdaptFormer offers the best trade-off between performance and efficiency across tasks. On a full temporal context video backbone, temporal modeling in the adapter adds training and inference cost without improving performance. 

\begin{table}[t]
\centering
\caption{\textbf{Comparison of PEFT Methods for Video.} Adapter methods perform best overall. Temporal modeling in the adapter adds training and inference cost without improving performance. Mean {\color{sdgray}$\pm$ sample std} over 5 seeds; \textbf{best} and \underline{second-best} are highlighted. \textsuperscript{\dag}Weights can be merged to obtain attentive probe throughput.}
\vspace{-0.5em}
\label{tab:peft_results}
\renewcommand{\arraystretch}{1.15}
\setlength{\tabcolsep}{5pt}

\begin{adjustbox}{max width=\textwidth}
\scriptsize
\begin{tabular}{
  @{}lll
  cc
  cc
  c
  cc
  c@{}
}
\toprule
& & & \multicolumn{2}{c}{\cellcolor{myblue}\textbf{Appearance}}
& \multicolumn{2}{c}{\cellcolor{myblue}\textbf{Motion}}
& &
\multicolumn{2}{c}{\cellcolor{myblue}\textbf{Spatial}} & \\
\cmidrule(lr){4-5}\cmidrule(lr){6-7}\cmidrule(lr){9-10}
& &
\makecell[l]{\textbf{Params}\\\textit{\scriptsize (M) $\downarrow$}}
& \makecell[c]{\textbf{CAER}\\\textit{\scriptsize top-1 $\uparrow$}}
& \makecell[c]{\textbf{NurViD}\\\textit{\scriptsize top-1 $\uparrow$}}
& \makecell[c]{\textbf{IndustReal}\\\textit{\scriptsize top-1 $\uparrow$}}
& \makecell[c]{\textbf{MammAlps}\\\textit{\scriptsize top-1 $\uparrow$}}
& \makecell[c]{\textbf{Clips/s}\\\textit{\scriptsize $\uparrow$}}
& \makecell[c]{\textbf{ScanNet}\\\textit{\scriptsize AbsRel $\downarrow$}}
& \makecell[c]{\textbf{VSPW}\\\textit{\scriptsize mIoU $\uparrow$}}
& \makecell[c]{\textbf{Frames/s}\\\textit{\scriptsize $\uparrow$}} \\
\midrule
 \multirow{2}{*}{\rotatebox[origin=c]{90}{\textbf{Base}}} &\g Full Finetuning & \g 87.1 & \g 60.3\sd{0.6} & \g \textbf{86.5}\sd{0.7} & \g \underline{67.2}\sd{0.5} & \g \textbf{71.7}\sd{0.7} & \g \underline{265.5} & \g 0.167\sd{0.006} & \g 40.2\sd{0.7} & \g \textbf{2041.3} \\
  &Attentive Probe & \textbf{7.4} & 59.7\sd{0.8} & 85.6\sd{0.3} & 62.9\sd{0.9} & 69.4\sd{0.8} & 264.9 & 0.160\sd{0.004} & 40.9\sd{0.2} & \underline{1860.5} \\
\midrule
 \multirow{3}{*}{\rotatebox[origin=c]{90}{\textbf{Selective}}} &\g BitFit & \g 7.4 & \g 60.4\sd{0.7} & \g 85.7\sd{1.0} & \g 63.7\sd{0.9} & \g 69.9\sd{1.3} & \g \textbf{265.7} & \g 0.153\sd{0.002} & \g 44.2\sd{0.5} & \g 1855.5 \\
  &LayerNorm & \underline{7.4} & 59.2\sd{1.0} & 85.7\sd{0.5} & 64.3\sd{0.3} & 69.3\sd{0.8} & 264.5 & 0.152\sd{0.003} & 44.4\sd{0.4} & 1859.2 \\
  &\g SSF & \g 7.6 & \g 60.2\sd{1.2} & \g 85.7\sd{0.7} & \g 65.4\sd{0.6} & \g 67.4\sd{2.0} & \g 217.4\textsuperscript{\dag} & \g \textbf{0.146}\sd{0.003} & \g 47.0\sd{0.2} & \g 1555.1\textsuperscript{\dag} \\
\midrule
 \multirow{3}{*}{\rotatebox[origin=c]{90}{\textbf{Additive}}} &VPT & 8.4 & 59.7\sd{1.1} & 85.7\sd{0.4} & 65.3\sd{0.3} & 70.0\sd{1.0} & 243.9 & 0.154\sd{0.004} & 47.8\sd{0.4} & 1649.6 \\
  &\g VFPT & \g 8.8 & \g 59.4\sd{0.7} & \g 85.8\sd{0.6} & \g 64.4\sd{0.3} & \g 67.2\sd{0.8} & \g 233.0 & \g 0.161\sd{0.005} & \g 47.5\sd{1.2} & \g 1640.2 \\
  &BPT & 8.5 & 60.2\sd{0.5} & 85.7\sd{0.6} & 65.0\sd{1.0} & 70.0\sd{1.4} & 244.3 & 0.156\sd{0.005} & 48.7\sd{0.4} & 1549.3 \\
\midrule
 \multirow{2}{*}{\rotatebox[origin=c]{90}{\textbf{Adapt}}} &\g AdaptFormer & \g 11.3 & \g 60.2\sd{1.3} & \g 86.0\sd{0.6} & \g 66.7\sd{0.3} & \g \underline{71.1}\sd{0.5} & \g 255.9 & \g 0.151\sd{0.004} & \g \textbf{49.9}\sd{0.5} & \g 1808.8 \\
  &ST-Adapter & 14.1 & \textbf{60.9}\sd{0.8} & \underline{86.2}\sd{0.8} & 67.0\sd{0.3} & 69.6\sd{0.9} & 231.7 & \underline{0.148}\sd{0.003} & \underline{49.2}\sd{0.5} & 1692.9 \\
\midrule
 \multirow{2}{*}{\rotatebox[origin=c]{90}{\textbf{Rank}}} &\g LoRA & \g 11.7 & \g \underline{60.7}\sd{1.0} & \g 86.0\sd{0.6} & \g \textbf{68.1}\sd{0.2} & \g 70.0\sd{2.5} & \g 244.4\textsuperscript{\dag} & \g 0.163\sd{0.010} & \g 47.3\sd{0.6} & \g 1728.7\textsuperscript{\dag} \\
  &DoRA & 13.3 & 59.6\sd{0.4} & 85.2\sd{0.4} & 66.2\sd{0.3} & 66.6\sd{1.1} & 237.4\textsuperscript{\dag} & 0.169\sd{0.007} & 45.6\sd{0.5} & 1683.0\textsuperscript{\dag} \\
\bottomrule
\end{tabular}
\end{adjustbox}
\vspace{-0.5em}
\end{table}

\vspace{-0.3em}
\subsection{Which Probe Works Best For Video?}
\begin{figure}
\vspace{-1em}
    \centering
    \includegraphics[width=0.97\textwidth]{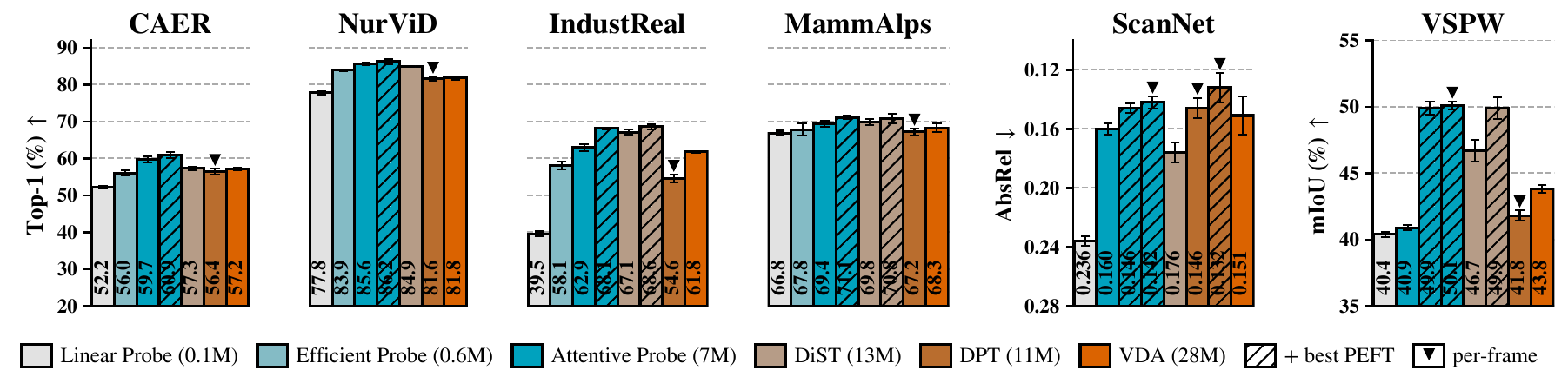}
    \vspace{-0.8em}
    \caption{\textbf{Comparison of Probe Methods for Video.} In isolation, the best probe varies per task, but combined with PEFT (hatched), a simple attentive probe outperforms or is within one standard deviation of task-specific probes. $\blacktriangledown$ Per-frame probes predict each frame's output using only that frame's features. Error bars show standard deviation over 5 seeds.}
    \vspace{-0.8em}
    \label{fig:probe_barplot} 
\end{figure}

\vspace{-0.4em}
\label{sec:exp_probe}
\textbf{Experimental Setup.} We compare six probing strategies on our ScarceVid benchmark, with results in Figure~\ref{fig:probe_barplot} and Appendix~\ref{app:decoder}. These include linear, efficient~\citep{psomas2026attention} and attentive probes~\citep{chen2024context,bardes2024revisiting}, as well as temporal-focused DiST~\citep{qing2023disentangling} and dense prediction probes DPT~\citep{ranftl2021vision} and VDA~\citep{chen2025video}. We test each probe in isolation and additionally combine the best PEFT method with the strongest probes per task.

\textbf{Results.} Alone, probe performance is strongly task-dependent, with no single method performing best across all tasks. With PEFT, these differences largely disappear: the generic attentive probe either outperforms or is within one standard deviation of the task-specific probes on all tasks. Adapting the backbone with PEFT thus removes the need to design task-specific probes.

Without PEFT, the attentive probe performs best for appearance tasks, suggesting that last-layer features already capture the required global information. For motion tasks and semantic segmentation (VSPW), DiST performs best, even beating the pixel-specialized DPT and VDA on VSPW. This points to the benefit of multi-layer aggregation for both. For depth estimation (ScanNet), DPT and VDA are the strongest, which can be explained by pixel-level prediction depending more heavily on early-layer features. DiST cannot match them here as its downsampling degrades spatial precision. Conversely, DPT and VDA underperform DiST on appearance and motion tasks. Among the lightweight probes, the linear probe is consistently weakest and serves mainly as a low-parameter lower bound. The efficient probe improves over it while still training an order of magnitude fewer parameters than attentive and task-specific probes, but does not scale to dense outputs.

Since spatial tasks predict every frame, they allow a direct comparison of probes with and without temporal modeling. The temporal modeling that VDA adds to DPT does not improve over the per-frame DPT on depth estimation, and the per-frame attentive probe shows the same trend on both spatial tasks. This suggests that temporal modeling in the probe is unnecessary for these tasks when the backbone already models time. Section~\ref{sec:exp_distribute} extends this analysis to all tasks.

\textbf{Conclusion.} In isolation, probe performance is task-dependent. When combined with PEFT, however, a generic attentive probe suffices, removing the need for task-specific probes.

\vspace{-0.3em}
\subsection{Do Input Frames or Temporal Modeling Matter More?} \label{sec:exp_uniform_input}
\vspace{-0.2em}

\textbf{Experimental Setup.} Having established effective probe and PEFT designs, we now examine how temporal context should be distributed across the model. To separate the effect of internal temporal modeling from additional input frames, we compare two regimes. The \emph{matched} setting ($T{=}T_{\text{backbone}}{=}T_{\text{peft}}{=}T_{\text{probe}}$; Section~\ref{sec:matched}) scales the input frames and all internal temporal contexts together. The \emph{input-fixed} setting $(T{\geq}T_{\text{backbone}}{=}T_{\text{peft}}{=}T_{\text{probe}})$ fixes the input clip at 16 frames while varying the temporal contexts of backbone, PEFT and probe jointly. The temporal context of each component must divide $T$ and cannot exceed it. When a component operates on a smaller context $T_c{<}T$, the component is applied independently to each $K{=}T/T_c$ segment and the outputs are recombined. We vary $T_c{\in}\{1, 2, 4, 8, 16\}$ for appearance and motion tasks and $T_c{\in}\{2, 4, 8, 16\}$ for spatial tasks, where the queries of the attentive probe are tubelets of two frames~\citep{carreira2024scaling,wang2025internvideo}. The performance within each task category is averaged. We use a ViT-B InternVideo-Next backbone, ST-Adapter as the temporal PEFT, and an attentive probe. We use ST-Adapter, as its temporal convolution allows the PEFT temporal context to be varied independently of the backbone. Throughput is reported in clips per second for classification tasks and frames per second for spatial tasks.

\begin{figure}
    \centering
    \includegraphics[width=\textwidth]{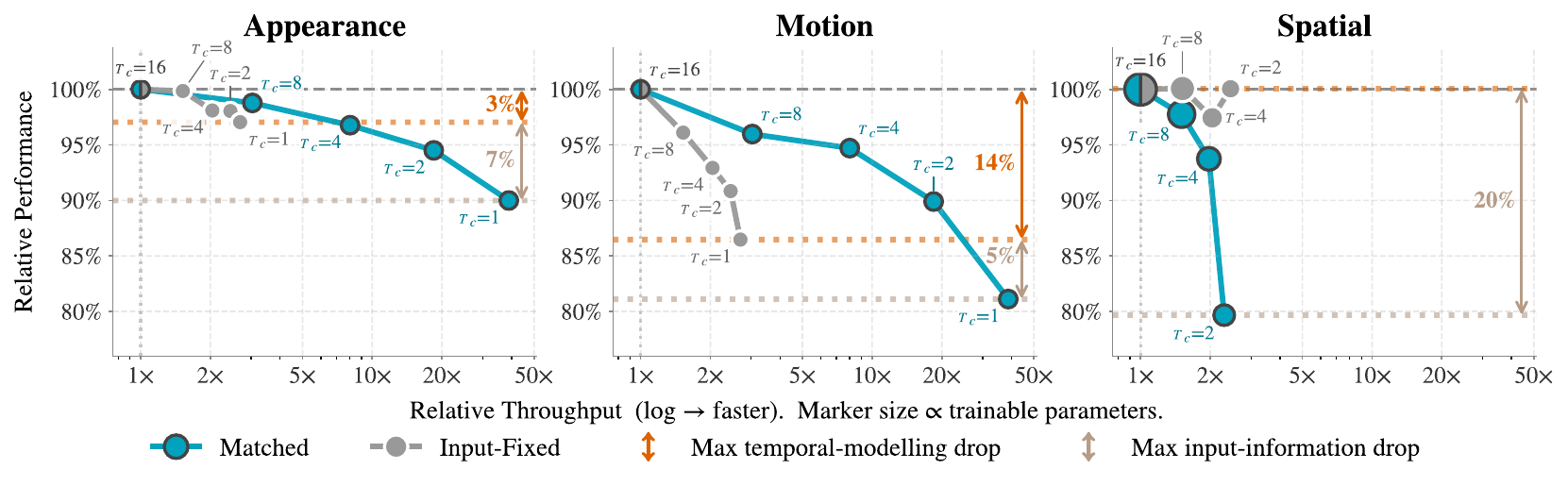}
    \vspace{-2em}
    \caption{\textbf{Disentangling Temporal Modeling from Input Frames.} Performance-throughput trade-offs when varying temporal context together with the input (matched) or only within the backbone, PEFT, and probe while keeping the input fixed at 16 frames (input-fixed). Motion tasks depend most on internal temporal modeling, while appearance and spatial tasks benefit most from more input frames. Performance is relative to $T{=}T_c{=}16$ and averaged per task category.}
    \label{fig:uniform_input}
    \vspace{-0.8em}
\end{figure}

\textbf{Results.} Figure~\ref{fig:uniform_input} shows the performance-throughput trade-offs for the matched and input-fixed settings. On appearance tasks, reducing only internal temporal modeling (input-fixed) gives a $3\%$ drop at $T_c{=}1$, while additionally reducing input frames (matched) reduces performance by another $7\%$. Motion tasks show the opposite: reducing internal temporal modeling accounts for most of the performance drop ($14\%$ out of $19\%$). The gap between matched and input-fixed settings falls below $1\%$ already at $T_c{=}2$, showing that once minimal temporal modeling is available, additional input frames provide little benefit and decrease throughput. On spatial tasks, the input-fixed setting stays near $T{=}T_c{=}16$ across the entire range, with $T_c{=}2$ providing substantially higher throughput with minimal accuracy loss. The observed $20\%$ drop therefore comes entirely from reduced input frames. These findings help explain why image backbones are competitive on appearance and spatial tasks (Section~\ref{sec:imagevsvideo}). For example, on appearance tasks, the input-fixed setting at $T_c{=}1$ linearly aggregates per-frame features and still recovers nearly all performance. This suggests that even video backbones rely little on temporal reasoning for these tasks. These trends are similar for VJEPA2 (Appendix~\ref{app:temporal_backbones}).

\textbf{Conclusion.} Appearance and spatial tasks benefit primarily from more input frames rather than temporal modeling. In contrast, motion tasks depend heavily on temporal reasoning.

\subsection{How to Distribute the Temporal Context?} \label{sec:exp_distribute}
\label{sec:exp_context}
\textbf{Experimental Setup.} Section~\ref{sec:exp_uniform_input} showed that the importance of internal temporal modeling varies strongly across task types, but used the same temporal context in backbone, PEFT, and probe. Here we ask how temporal context is most effectively distributed. As well as the input-fixed setting with $T{=}16$ and $T_{\text{backbone}}{=}T_{\text{peft}}{=}T_{\text{probe}}{=}T_c$, we evaluate three variants where the temporal context of one model component is fixed at 16 while the other two vary with $T_c$. For instance, fixing the backbone gives $T{=}T_{\text{backbone}}{=}16$ and $T_{\text{peft}}{=}T_{\text{probe}}{=}T_c$. Setup otherwise matches Section~\ref{sec:exp_uniform_input}.

\begin{figure}[t]
    \centering
    \includegraphics[width=\textwidth]{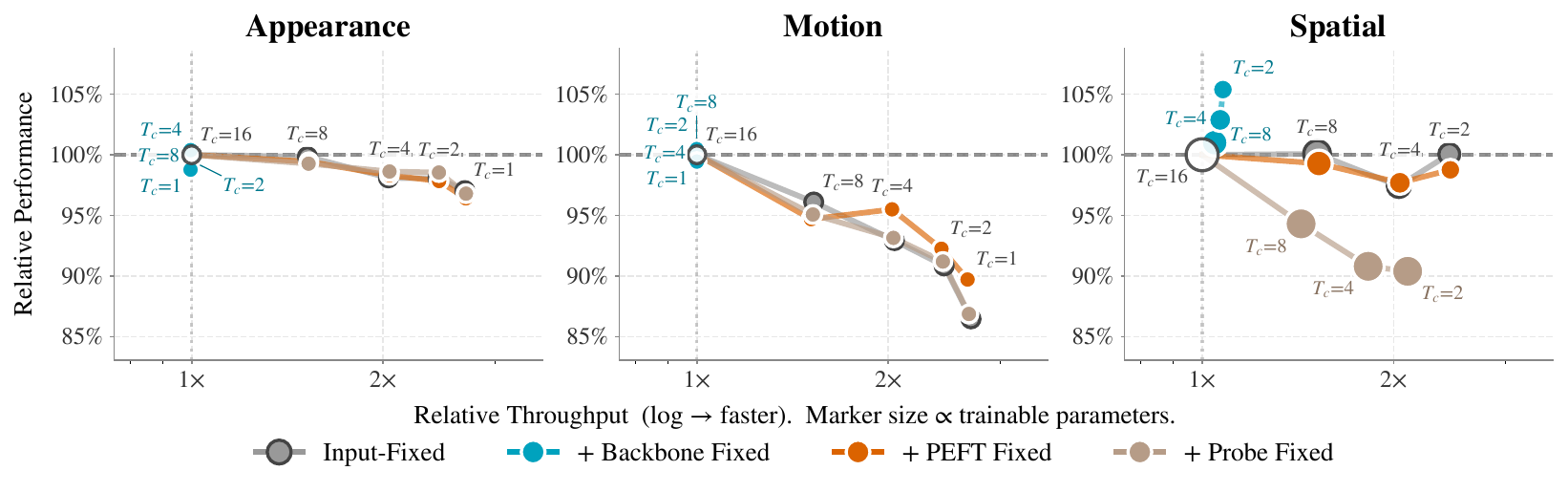}
    \vspace{-2em}
    \caption{\textbf{How to Distribute Temporal Context.} Input-fixed varies the temporal context $T_c$ of backbone, PEFT, and probe jointly, while each ``+X Fixed'' curve keeps component X at 16 frames and varies the other two with $T_c$. A curve above input-fixed thus means that component X benefits from temporal context. Temporal context is most effective in the backbone, and for spatial tasks harmful in the probe. Performance is relative to $T{=}T_c{=}16$ and averaged per task category.}
    \label{fig:comp_modules}
    \vspace{-0.8em}
\end{figure}

\textbf{Results.} Figure~\ref{fig:comp_modules} compares the four temporal context allocations. On appearance tasks, all configurations stay within $\sim$4\% of the full temporal-context reference ($T{=}T_{\text{backbone}}{=}T_{\text{peft}}{=}T_{\text{probe}}$), again showing that internal temporal modeling contributes little to these tasks. Throughput, however, varies substantially: reducing backbone context yields the largest speedup, while keeping $T_{\text{backbone}}{=}16$ leaves throughput nearly unchanged regardless of PEFT or probe context. Between PEFT and probe, the difference in contribution is marginal and varies per dataset. On motion tasks, temporal modeling is dominated by the backbone. Keeping $T_{\text{backbone}}{=}16$ maintains performance close to the reference regardless of $T_c$, while reducing it drops performance sharply. When $T_{\text{backbone}}{<}T$, allocating more temporal context to the PEFT modules is more effective than allocating it to the probe. This \emph{+PEFT Fixed} setting corresponds to a long PEFT context on a short backbone context (Section~\ref{sec:method_decoupled_peft}), where the temporal convolution of ST-Adapter mixes information across independently processed backbone segments. However, when computation permits, increasing temporal context in the backbone remains most effective. Spatial tasks show a different pattern. With $T_{\text{backbone}}{=}16$, performance \emph{improves} as temporal context in PEFT and probe shrinks. The fixed probe setting confirms the probe is the main source of degradation: when the probe operates over many frames, performance drops regardless of the rest of the model. This is consistent with Section~\ref{sec:exp_probe}, where the temporal modeling of VDA does not improve over the per-frame DPT on depth estimation. The same degradation on spatial tasks also occurs for VJEPA2 and (image-pretrained) DINOv3 (Appendix~\ref{app:temporal_backbones}). We attribute this to dense prediction becoming harder when the low-capacity probe must integrate across many temporal positions, trading spatial precision for unnecessary temporal aggregation. This also reveals why uniformly increasing temporal context showed little benefit in Section~\ref{sec:exp_uniform_input}: gains from temporal modeling in the backbone are canceled by the degradation from the temporal probe. For spatial tasks, temporal context is most effective in the backbone.

\vspace{1em}
\textbf{Conclusion.} Temporal context should be allocated selectively rather than uniformly. For appearance, throughput is best saved by reducing backbone context. Motion benefits most from temporal context in the backbone, though allocating it to PEFT can improve performance when high throughput is required. Spatial tasks also favor temporal context in the backbone, but \emph{not} in the probe.

\vspace{-0.5em}
\section{Discussion}
\label{sec:conclusion}
\vspace{-0.5em}
\noindent\textbf{Conclusion.} 
We presented a systematic study of model adaptation for low-resource video understanding across appearance, motion and spatial tasks. We benchmarked a diverse set of PEFT and probe approaches and analyzed how temporal context at the backbone, PEFT, and probe levels affects adaptation. Our experiments provide insight into when different adaptation strategies succeed and highlight temporal context allocation as a key consideration for efficient video adaptation.

\noindent\textbf{Recommendations.} 
Our experiments suggest the best adaptation strategy depends strongly on the type of video task. For appearance tasks, a video-pretrained model (InternVideo-Next) outperforms image-pretrained alternatives, but only lightweight adaptation is required. In particular, image-style adapter PEFT methods, combined with limited temporal context in the backbone (\textit{e.g.} $T_{\text{backbone}}{=}2$) and the probe (\textit{e.g.} $T_{\text{probe}}{=}2$), provide a good efficiency-accuracy trade-off.
For motion tasks, temporal modeling in the backbone is critical. Here, video-pretrained backbones are more effective than image models and increasing temporal reasoning in the backbone leads to the largest performance gains. In contrast, temporal modeling in the PEFT and probe is not necessary, although it can partially compensate for reduced backbone context when higher throughput is required.
Spatial tasks are different again. While adapting DINOv3 is most effective, likely due to large-scale dense pretraining, we find benefits from temporal modeling within video backbones. However, when adapting with limited data, low-capacity PEFT and probes struggle to preserve spatial detail while integrating temporal context. This means temporal context in PEFT and \emph{especially} the probe should be minimized.

\noindent\textbf{Limitations.} While each task category contains two datasets chosen to differ in domain, viewpoint, and annotation granularity, additional datasets would provide broader support for category-level conclusions. Furthermore, our study focuses on ViT backbones, which underpin most current vision foundation models, but we have not verified whether our findings transfer to other architectures or to video LLMs. Finally, our temporal context analysis varies one or two components at a time, as the temporal context search space over backbone, per-layer PEFT, and probe is combinatorially large. These choices likely interact, and an efficient search over the joint space is left to future work.

\noindent\textbf{Broader Impact.} Efficient adaptation of video foundation models can make advanced video understanding systems more accessible by reducing the computation and annotation costs required for deployment. This is particularly relevant for applications where labeled video data is scarce or expensive, such as healthcare, scientific imaging, manufacturing, robotics and environmental monitoring. By studying parameter-efficient adaptation in low-resource settings our work may help enable video models in settings where large-scale training or finetuning is impractical.

\begin{ack}
    This work is supported by the Dutch Research Council (NWO) under a Veni grant (VI.Veni.222.160).
\end{ack}

\bibliographystyle{plainnat}
\bibliography{main}

\appendix

\section{Experiment Details}
\label{app:experiment_details}

\subsection{Backbone}
All experiments use an InternVideo-Next~\citep{wang2025internvideo} ViT-B backbone with patch size 14 and input resolution $224\times224$, which yields $\frac{224}{14}\times\frac{224}{14}=256$ spatial tokens per frame, unless stated otherwise. To keep the token budget consistent across image and video backbones, we fix the total number of tokens to $256\times8=2048$, corresponding to 8 input frames for models that tokenize each frame separately (InternVideo-Next, SigLIP2, DINOv3). For VJEPA2~\citep{assran2025v} that processes spatio-temporal tubelets of size 2, we use 16 input frames resulting in $16/2=8$ temporal positions. The temporal context experiments in Sections~\ref{sec:exp_uniform_input} and~\ref{sec:exp_distribute} instead use 16 input frames for all backbones. Image models do not produce temporal positional encodings; we therefore add learnable positional embeddings to their output features before the probe head as in \citet{simeoni2025dinov3}.

\subsection{Parameter-Efficient Finetuning Methods}

\paragraph{Selective methods.}
\textit{BitFit}~\citep{zaken2022bitfit} unfreezes only the bias terms of every linear layer in the backbone. \textit{LayerNorm} tunes only the scale and shift parameters of every LayerNorm in the backbone. \textit{SSF}~\citep{lian2022scaling} inserts a learnable per-channel affine transformation $\hat{x} = \gamma \odot x + \beta$ after each of the QKV projection, attention output projection, and MLP activations. Scale $\gamma$ is initialized $\sim\mathcal{N}(1,\,0.02)$ and bias $\beta\sim\mathcal{N}(0,\,0.02)$.

\paragraph{Additive methods.}
All prompt methods use the deep variant, inserting tokens at every transformer layer. The tokens are initialized $\sim\mathcal{N}(0,\,0.02)$ and discarded after each block.
We sweep the number of prompt tokens $N_p\in\{16,64,256\}$ for all three methods. \textit{VPT}~\citep{jia2022visual} prepends $N_p$ learnable embedding vectors to the token sequence at each layer. \textit{VFPT} extends VPT by replacing the first $\lfloor N_p/2 \rfloor$ prompt tokens at each layer with the real part of their 2D discrete Fourier transform (over the token and channel axes), encouraging frequency-diverse structure in the prompt representation. \textit{BPT} parameterizes the $N_p$ prompt tokens at each layer as the product of two low-rank matrices $\mathbf{P}=\mathbf{A}\mathbf{B}^\top$. We set the bottleneck rank to 100.

\paragraph{Adapter methods.}
\textit{AdaptFormer}~\citep{chen2022adaptformer} inserts a bottleneck MLP (down-projection $\to$ ReLU $\to$ up-projection) in \emph{parallel} with the MLP block of every transformer layer. The up-projection is zero-initialized and its output is scaled by a learnable scalar initialized to $10^{-3}$, so the adapter is identity at the start of training. We sweep bottleneck rank $r \in \{64, 128, 256\}$ with dropout $0.0$.

\textit{ST-Adapter}~\citep{pan2022st} is placed \emph{sequentially} before both the attention and the MLP sub-layers in each block. It projects tokens to a bottleneck of width $r$, applies a depthwise 3D convolution with kernel $3{\times}1{\times}1$ to mix information along the temporal axis while preserving spatial resolution, and projects back to the original dimension. The convolution is zero-initialized so the adapter begins as an identity; there is no intermediate activation. We sweep $r\in\{64, 128, 256\}$ with dropout $0.0$; in the ablations of Sections~\ref{sec:exp_uniform_input} and~\ref{sec:exp_distribute} we fix $r{=}128$ to reduce compute.

\paragraph{Reparameterization methods.}
\textit{LoRA}~\citep{hu2022lora} adds a low-rank residual $\Delta\mathbf{W}=\mathbf{B}\mathbf{A}$ to the QKV weight of every attention block, where $\mathbf{A}\in\mathbb{R}^{r\times d}$ (Kaiming initialized) and $\mathbf{B}\in\mathbb{R}^{3d\times r}$ (zero initialized), so the adapter is identity at initialization. We sweep rank $r\in\{64,128,256\}$ with dropout 0.1 applied to the input of $\mathbf{B}$.

\textit{DoRA}~\citep{liu2024dora} extends LoRA by decomposing the effective weight into a direction component and a learnable magnitude vector $\mathbf{m}$, where $\mathbf{m}$ is initialized to the row-wise $\ell_2$ norms of the pretrained QKV weight, so DoRA is also identical to the pretrained model at initialization. Other settings match LoRA.

\subsection{Probe}
\paragraph{Linear probe.}
This simple baseline applies global average pooling over all backbone tokens (spatial and temporal) and maps the resulting vector to class logits with a single linear layer. This probe is also used for full finetuning.

\paragraph{Attentive probe.}
For depth estimation and semantic segmentation we adopt the attentive probe of \citet{carreira2024scaling}, which inserts a single cross-attention block (with a subsequent MLP; $12$ heads, MLP ratio $4.0$) between a set of learnable spatial queries and the backbone tokens. We set the probe patch size to $8$, yielding $\frac{224}{8}\times\frac{224}{8} = 784$ spatial query tokens per frame; for a clip of $T$ frames the queries are tiled as $\frac{T}{d_t}\times784$ where $d_t{=}2$ is the query tubelet size. For the per-frame variant in Figure~\ref{fig:probe_barplot}, we use $d_t{=}1$ and restrict the queries of each frame to attend only to that frame's tokens. For classification we use a single learnable query that cross-attends to all backbone tokens, producing a global representation. In both cases a linear layer projects the query outputs to the target dimension.

\paragraph{Efficient probe.}
Following \citet{psomas2026attention}, we use a single-head cross-attention pooler with $N_q{=}32$ learnable queries. Each query attends to the full set of backbone tokens via an efficient value-only projection. The aggregated vectors are then fed to a linear head. The efficient probe is not evaluated on the spatial tasks because its design does not transfer to dense prediction. Extending the attentive probe's one learnable query per $8{\times}8{\times}2$ output tubelet for dense tasks gives $\frac{T}{2}\times(\frac{224}{8})^2$ queries for a clip of $T$ frames. Thus the efficient probe would require each of the thousands of spatial query positions to have its own set of 32 queries. This vastly increases the attention cost and removes exactly what makes the probe efficient. 

\paragraph{DiST.}
Following \citet{qing2023disentangling}, we fuse the input frames encoded by a lightweight temporal encoder with the intermediate features from all backbone layers through a spatial-temporal integration network. The lightweight temporal encoder is a 3D CNN with kernel $5{\times}14{\times}14$, stride $1{\times}14{\times}14$, and extracts $96$-dimensional temporal features from the input frames. The backbone features are first linearly projected to the shared integration dimension of $384$. The integration network is an MLP with MLP ratio $1.0$, spatial-temporal MLP ratio $0.25$, and temporal kernel size $3$. The final output is a sequence of spatio-temporal features of dimension $384$. 

Since \citet{qing2023disentangling} consider only action recognition, where a single pooled representation suffices, we extend DiST to dense prediction by replacing the original pooling step with an attentive probe that cross-attends between the output features and the learnable queries.

\paragraph{DPT.}
Following \citet{ranftl2021vision}, DPT gets the dense predictions from four backbone layers sampled at uniform depth fractions $k/4$ for $k{=}1,2,3,4$ (layers $3, 6, 9, 12$ for a 12-layer backbone). Each layer's token map is projected with a $1{\times}1$ convolution to intermediate channels $[96,192,384,768]$, then spatially resized and fused top-down through residual fusion blocks. The fused map is processed by a two-stage convolutional head that produces per-pixel logits directly, without any additional linear projection on token representations. For depth estimation the output is scaled to $[0, d_{\max}]$ via a sigmoid.

\paragraph{VDA.}
Video Depth Anything~\citep{chen2025video} extends DPT with temporal self-attention at four intermediate decoder stages. Each temporal module contains one transformer block with two temporal self-attention sub-layers ($8$ heads, absolute sinusoidal positional encodings along the time axis) and zero-initialized output projections, so the model initializes identically to a per-frame DPT. 

\subsection{Training}
\textbf{Optimization.} All models are trained with AdamW ($\beta_1{=}0.9$, $\beta_2{=}0.999$, $\epsilon{=}10^{-8}$, weight decay $0.01$) for 4{,}000 steps. We use a linear warmup over the first 400 steps (warmup factor 0.1) followed by cosine decay to zero, with a global batch size of 32 on 2$\times$H100 GPUs. For classification we sweep learning rates over $\{10^{-4}, 2.5 \times 10^{-4}, 5 \times 10^{-4}, 10^{-3}\}$. For depth estimation and semantic segmentation we use a higher range of $\{5 \times 10^{-4}, 10^{-3}, 2.5 \times 10^{-3}, 5 \times 10^{-3}\}$. The probe head uses drop-path regularization with rate 0.1.

\textbf{Augmentation.} We use minimal data augmentation during training. Consistent with findings from image-based PEFT benchmarks~\citep{mai2025lessons}, we found that strategies such as RandAugment~\citep{cubuk2020randaugment}, mix-up~\citep{zhang2017mixup}, and random erasing~\citep{zhong2020random} hurt performance, especially on smaller datasets. For all tasks we apply only random resized cropping, with scale in $[0.3, 1.0]$ and aspect ratio in $[0.75, 1.33]$.

\textbf{Frame sampling.} For classification we use uniform segment sampling: the video is split into $T$ equal-length segments and one frame is drawn uniformly at random from each, giving global coverage of the action. For depth estimation and segmentation we instead use dense sampling of $T$ consecutive frames, preserving the local temporal structure needed for per-frame dense prediction.

\subsection{Evaluation}
\textbf{Classification.} We follow the standard three-crop protocol, averaging softmax scores across the three spatial crops and reporting top-1 accuracy.

\textbf{Dense prediction.} At test time, each video is partitioned into non-overlapping consecutive chunks of $T$ frames that together cover the full sequence. Each chunk is processed independently with a single center crop, yielding per-frame predictions for every frame. Metrics (AbsRel for depth, mIoU for segmentation) are computed per frame across all chunks and averaged over the entire test set, giving true per-frame dataset-level scores.

\textbf{Efficiency metrics.}
Inference throughput is measured using 2 NVIDIA H100 GPUs on the test set with a batch size of 16 clips per GPU. We synchronize the GPU before and after each forward so that data loading and pre-/post-processing are excluded. Throughput is the total number of processed units summed over both GPUs, divided by the forward time of the slowest GPU. For classification we report clips per second since a single prediction is made per clip. For depth estimation and semantic segmentation a prediction is made for every frame, so we report frames per second. Peak memory is the maximum per-GPU memory allocated during inference on the test set. Peak memory is averaged over all datasets, clips per second over the four classification datasets, and frames per second over the two spatial datasets. The number of trainable parameters varies per dataset, because the probe depends on the number of classes, output type, PEFT rank, and number of prompts. We therefore report the mean over all datasets. 

\section{Additional Results}

\subsection{Backbone size} \label{app:backbone_size}
In Table~\ref{tab:vitb_vs_vitl} we report a backbone size comparison for InternVideo-Next. ViT-B with AdaptFormer outperforms ViT-L with a linear probe on every task. When the same adaptation method is applied to both backbones, ViT-L is better on all 24/24 comparisons, with the largest gains on motion and spatial tasks. Importantly, gains from different PEFT and probing methods remain similar, regardless of the backbone size used.

\begin{table}[h]
\centering
\caption{\textbf{ViT-B vs.\ ViT-L for InternVideo-Next.} ViT-B with AdaptFormer beats ViT-L with a linear probe. When the same method is used on both backbones, ViT-L outperforms on all 24/24 comparisons, with the largest gains on motion and spatial tasks. }
\label{tab:vitb_vs_vitl}
\renewcommand{\arraystretch}{1.15}
\setlength{\tabcolsep}{5pt}
\begin{adjustbox}{max width=\textwidth}
\scriptsize
\begin{tabular}{
  @{}ll
  cc
  cc
  cc@{}
}
\toprule
& & \multicolumn{2}{c}{\cellcolor{myblue}\textbf{Appearance}}
& \multicolumn{2}{c}{\cellcolor{myblue}\textbf{Motion}}
& \multicolumn{2}{c}{\cellcolor{myblue}\textbf{Spatial}} \\
\cmidrule(lr){3-4}\cmidrule(lr){5-6}\cmidrule(lr){7-8}
\textbf{Method} & \textbf{Backbone}
& \makecell[c]{\textbf{CAER}\\\textit{\scriptsize top-1 $\uparrow$}}
& \makecell[c]{\textbf{NurViD}\\\textit{\scriptsize top-1 $\uparrow$}}
& \makecell[c]{\textbf{IndustReal}\\\textit{\scriptsize top-1 $\uparrow$}}
& \makecell[c]{\textbf{MammAlps}\\\textit{\scriptsize top-1 $\uparrow$}}
& \makecell[c]{\textbf{ScanNet}\\\textit{\scriptsize AbsRel $\downarrow$}}
& \makecell[c]{\textbf{VSPW}\\\textit{\scriptsize mIoU $\uparrow$}} \\
\midrule
\multirow{2}{*}{Linear Probe}
 & \g ViT-B & \g 52.2\sd{0.4} & \g 77.8\sd{0.5} & \g 39.5\sd{0.7} & \g 66.8\sd{0.7} & \g 0.236\sd{0.003} & \g 40.4\sd{0.2}\\
 & ViT-L    & \textbf{55.0} & \textbf{83.5} & \textbf{41.4} & \textbf{70.7} & \textbf{0.232} & \textbf{42.1} \\
\midrule
\multirow{2}{*}{Attentive Probe}
 & \g ViT-B & \g 59.7\sd{0.8} & \g 85.6\sd{0.3} & \g 62.9\sd{0.9} & \g 69.4\sd{0.8} & \g 0.160\sd{0.004} & \g 40.9\sd{0.2} \\
 & ViT-L    & \textbf{62.7} & \textbf{88.5} & \textbf{67.0} & \textbf{74.0} & \textbf{0.135} & \textbf{47.3} \\
\midrule
\multirow{2}{*}{LoRA}
 & \g ViT-B & \g 60.7\sd{1.0} & \g 86.0\sd{0.6} & \g 68.1\sd{0.2} & \g 70.0\sd{2.5} & \g 0.163\sd{0.010} & \g 47.3\sd{0.6} \\
 & ViT-L    & \textbf{63.1} & \textbf{88.5} & \textbf{72.8} & \textbf{75.9} & \textbf{0.142} & \textbf{55.2} \\
\midrule
\multirow{2}{*}{AdaptFormer}
 & \g ViT-B & \g 60.2\sd{1.3} & \g 86.0\sd{0.6} & \g 66.7\sd{0.3} & \g 71.1\sd{0.5} & \g 0.151\sd{0.004} & \g 49.9\sd{0.5} \\
 & ViT-L    & \textbf{63.4} & \textbf{88.6} & \textbf{70.0} & \textbf{76.3} & \textbf{0.118} & \textbf{57.6} \\
\bottomrule
\end{tabular}
\end{adjustbox}
\end{table}

\subsection{Probe} \label{app:decoder}
Table~\ref{tab:decoder_results} reports the numerical values corresponding to the probe comparison visualized in Figure~\ref{fig:probe_barplot}. We evaluate six probe variants spanning lightweight (linear, efficient, attentive) and heavier task-specific decoders (DiST~\citep{qing2023disentangling}, DPT~\citep{ranftl2021vision}, VDA~\citep{chen2025video}) across all six tasks. 

\begin{table}[h]
\centering
\caption{\textbf{Comparison of Probe Methods for InternVideo-Next ViT-B.} In isolation, the best probe varies per task. The efficient probe is not applicable to dense prediction tasks. Mean $\pm$ sample std over 5 seeds.}
\label{tab:decoder_results}
\renewcommand{\arraystretch}{1.15}
\setlength{\tabcolsep}{5pt}

\begin{adjustbox}{max width=\textwidth}
\scriptsize
\begin{tabular}{
  @{}ll
  cc
  cc
  c
  cc
  c@{}
}
\toprule
& & \multicolumn{2}{c}{\cellcolor{myblue}\textbf{Appearance}}
& \multicolumn{2}{c}{\cellcolor{myblue}\textbf{Motion}}
& &
\multicolumn{2}{c}{\cellcolor{myblue}\textbf{Spatial}} & \\
\cmidrule(lr){3-4}\cmidrule(lr){5-6}\cmidrule(lr){8-9}
& \makecell[l]{\textbf{Params}\\\textit{\scriptsize (M) $\downarrow$}}
 & \makecell[c]{\textbf{CAER}\\\textit{\scriptsize top-1 $\uparrow$}}
& \makecell[c]{\textbf{NurViD}\\\textit{\scriptsize top-1 $\uparrow$}}
& \makecell[c]{\textbf{IndustReal}\\\textit{\scriptsize top-1 $\uparrow$}}
& \makecell[c]{\textbf{MammAlps}\\\textit{\scriptsize top-1 $\uparrow$}}
& \makecell[c]{\textbf{Clips/s}\\\textit{\scriptsize $\uparrow$}}
& \makecell[c]{\textbf{ScanNet}\\\textit{\scriptsize AbsRel $\downarrow$} / \textit{\scriptsize$\delta_1$} $\uparrow$}
& \makecell[c]{\textbf{VSPW}\\\textit{\scriptsize mIoU $\uparrow$}}
& \makecell[c]{\textbf{Frames/s}\\\textit{\scriptsize $\uparrow$}} \\
\midrule

  \g Linear Probe & \g 0.1   & \g 52.2\sd{0.4} & \g 77.8\sd{0.5} & \g 39.5\sd{0.7} & \g 66.8\sd{0.7} & \g 266.3 & \g 0.236\sd{0.003} / 62.7\sd{0.2} & \g 40.4\sd{0.2} & \g 2046.8   \\
  Efficient Probe&  0.6 &  56.0\sd{0.7} &  83.9\sd{0.2} &  58.1\sd{1.1} &  67.8\sd{1.7} & 263.9 &  NA &  NA & NA   \\
 \g Attentive Probe  &\g 7.4  &\g 59.7\sd{0.8} &\g 85.6\sd{0.3} &\g 62.9\sd{0.9} &\g 69.4\sd{0.8} &\g 264.9 &\g 0.160\sd{0.004} / 79.1\sd{0.5} &\g 40.9\sd{0.2} &\g 1860.5   \\

   DiST & 12.9 & 57.3\sd{0.4} & 84.9\sd{0.2} & 67.1\sd{0.7} & 69.8\sd{0.8} & 225.2 & 0.176\sd{0.007} / 76.7\sd{1.2} & 46.7\sd{0.8} & 1630.7   \\
 \g DPT   & \g 10.6       & \g 56.4\sd{0.8} & \g 81.6\sd{0.6} & \g 54.6\sd{1.0} & \g 67.2\sd{0.9} & \g 195.8 & \g 0.146\sd{0.007} / 81.4\sd{1.0} & \g 41.8\sd{0.4} & \g 1451.3   \\
   VDA    & 27.6      & 57.2\sd{0.3} & 81.8\sd{0.5} & 61.8\sd{0.2} & 68.3\sd{1.1} & 182.1 & 0.151\sd{0.013} / 80.5\sd{3.5} & 43.8\sd{0.3} & 1353.1  \\

\bottomrule
\end{tabular}
\end{adjustbox}
\end{table}

\subsection{Temporal Context Analysis with Additional Backbones}
\label{app:temporal_backbones}

Our temporal context analysis in Sections~\ref{sec:exp_uniform_input} and~\ref{sec:exp_context} uses an InternVideo-Next ViT-B backbone. To verify that the observed trends are not specific to a single backbone, we repeat both experiments with VJEPA2 ViT-L under an identical protocol. We additionally extend the temporal context distribution analysis of Section~\ref{sec:exp_context} to DINOv3 ViT-B on the spatial tasks, since that is where DINOv3 outperforms the video backbones in Table~\ref{tab:backbones}.

\noindent \textbf{VJEPA2 extension.}
As in Section~\ref{sec:exp_uniform_input}, Table~\ref{tab:rebuttal_vjepa2_fig3} separates the effects of temporal modeling and input information. We compare a \textit{reference} setting, which uses a temporal context of 16 frames in all components, with two reduced temporal context settings. \textit{Matched} reduces both the input and temporal context to VJEPA2's minimum of 2 frames while \textit{input-fixed} retains the input of 16 frames but processes them as independent two-frame segments in the backbone, PEFT, and probe. Max temporal-modeling drop is the loss from reducing only internal temporal context (Reference $\rightarrow$ Input-Fixed); max input-information drop is the additional loss from also reducing input frames (Input-Fixed $\rightarrow$ Matched).

\begin{table}[H]
\centering
\caption{\textbf{Disentangling Temporal Modeling from Input Frames with InternVideo-Next and VJEPA2.} Both use ST-Adapter and an attentive probe. Reference uses $T_c{=}16$ in all components; Matched and Input-Fixed use $T_c{=}2$ (VJEPA2's tubelet size). Max temporal-modeling drop: Reference $\to$ Input-Fixed; max input-information drop: Input-Fixed $\to$ Matched. Relative drop in parentheses.}
\label{tab:rebuttal_vjepa2_fig3}
\renewcommand{\arraystretch}{1.15}
\setlength{\tabcolsep}{5pt}
\begin{adjustbox}{max width=\textwidth}
\scriptsize
\begin{tabular}{
  @{}l
  cc
  cc
  cc@{}
}
\toprule
& \multicolumn{2}{c}{\cellcolor{myblue}\textbf{Appearance}}
& \multicolumn{2}{c}{\cellcolor{myblue}\textbf{Motion}}
& \multicolumn{2}{c}{\cellcolor{myblue}\textbf{Spatial}} \\
\cmidrule(lr){2-3}\cmidrule(lr){4-5}\cmidrule(lr){6-7}
& \makecell[c]{\textbf{CAER}\\\textit{\scriptsize top-1 $\uparrow$}}
& \makecell[c]{\textbf{NurViD}\\\textit{\scriptsize top-1 $\uparrow$}}
& \makecell[c]{\textbf{IndustReal}\\\textit{\scriptsize top-1 $\uparrow$}}
& \makecell[c]{\textbf{MammAlps}\\\textit{\scriptsize top-1 $\uparrow$}}
& \makecell[c]{\textbf{ScanNet}\\\textit{\scriptsize AbsRel $\downarrow$}}
& \makecell[c]{\textbf{VSPW}\\\textit{\scriptsize mIoU $\uparrow$}} \\
\midrule
\textbf{InternVideo-Next ViT-B}\\
Reference ($T_c{=}16$)          & 62.4 & 85.7 & 68.8 & 72.9 & 0.140 & 49.5 \\
Matched ($T_c{=}2$)             & 56.0 & 85.1 & 59.9 & 67.6 & 0.195 & 48.6 \\
Input-Fixed ($T_c{=}2$)         & 60.3 & 85.2 & 60.3 & 68.5 & 0.143 & 50.5 \\
\midrule
Max temporal-modeling drop     & 2.1 \scriptsize{(3.3\%)} & 0.5 \scriptsize{(0.6\%)} & 8.5 \scriptsize{(12.3\%)} & 4.4 \scriptsize{(6.1\%)} & 0.003 \scriptsize{(1.9\%)} & $-$1.0 \scriptsize{($-$2.0\%)} \\
Max input-information drop      & 4.4 \scriptsize{(7.0\%)} & 0.1 \scriptsize{(0.1\%)} & 0.4 \scriptsize{(0.7\%)} & 0.9 \scriptsize{(1.2\%)} & 0.052 \scriptsize{(37.1\%)} & 1.8 \scriptsize{(3.7\%)} \\
\midrule
\textbf{VJEPA2 ViT-L}\\
Reference ($T_c{=}16$)          & 58.1 & 84.0 & 66.3 & 75.2 & 0.139 & 46.9 \\
Matched ($T_c{=}2$)             & 50.4 & 82.2 & 52.8 & 71.0 & 0.150 & 43.6 \\
Input-Fixed ($T_c{=}2$)         & 56.1 & 83.2 & 58.4 & 72.8 & 0.142 & 44.5 \\
\midrule
Max temporal-modeling drop     & 2.0 \scriptsize{(3.5\%)} & 0.8 \scriptsize{(1.0\%)} & 7.9 \scriptsize{(12.0\%)} & 2.3 \scriptsize{(3.1\%)} & 0.003 \scriptsize{(2.0\%)} & 2.4 \scriptsize{(5.2\%)} \\
Max input-information drop      & 5.7 \scriptsize{(9.8\%)} & 1.0 \scriptsize{(1.2\%)} & 5.6 \scriptsize{(8.5\%)} & 1.8 \scriptsize{(2.5\%)} & 0.008 \scriptsize{(6.1\%)} & 0.9 \scriptsize{(1.9\%)} \\
\bottomrule
\end{tabular}
\end{adjustbox}
\end{table}

As in Section~\ref{sec:exp_context}, Table~\ref{tab:rebuttal_vjepa2_fig4} examines where temporal context should be allocated. Starting from \textit{input-fixed}, where all components process independent two-frame segments, we allow either the backbone, PEFT module, or probe to jointly process all 16 frames while keeping the remaining components at two frames. This isolates the benefit of long temporal context within each component.

\begin{table}[h]
\centering
\caption{\textbf{How to Distribute Temporal Context with InternVideo-Next and VJEPA2.} Both use ST-Adapter and an attentive probe. Input-Fixed varies backbone, PEFT, and probe context jointly at $T_c{=}2$; each ``+X Fixed'' variant instead keeps one component at $T{=}16$ while the other two shrink to $T_c{=}2$. Reference is $T{=}T_{\text{backbone}}{=}T_{\text{peft}}{=}T_{\text{probe}}{=}16$.}
\label{tab:rebuttal_vjepa2_fig4}
\renewcommand{\arraystretch}{1.15}
\setlength{\tabcolsep}{5pt}
\begin{adjustbox}{max width=\textwidth}
\scriptsize
\begin{tabular}{
  @{}l
  cc
  cc
  cc@{}
}
\toprule
& \multicolumn{2}{c}{\cellcolor{myblue}\textbf{Appearance}}
& \multicolumn{2}{c}{\cellcolor{myblue}\textbf{Motion}}
& \multicolumn{2}{c}{\cellcolor{myblue}\textbf{Spatial}} \\
\cmidrule(lr){2-3}\cmidrule(lr){4-5}\cmidrule(lr){6-7}
& \makecell[c]{\textbf{CAER}\\\textit{\scriptsize top-1 $\uparrow$}}
& \makecell[c]{\textbf{NurViD}\\\textit{\scriptsize top-1 $\uparrow$}}
& \makecell[c]{\textbf{IndustReal}\\\textit{\scriptsize top-1 $\uparrow$}}
& \makecell[c]{\textbf{MammAlps}\\\textit{\scriptsize top-1 $\uparrow$}}
& \makecell[c]{\textbf{ScanNet}\\\textit{\scriptsize AbsRel $\downarrow$}}
& \makecell[c]{\textbf{VSPW}\\\textit{\scriptsize mIoU $\uparrow$}} \\
\midrule
\textbf{InternVideo-Next ViT-B}\\
Reference ($T{=}16$)            & 62.4 & 85.7 & 68.8 & 72.9 & 0.140 & 49.5 \\
Input-Fixed ($T_c{=}2$)         & 60.3 & 85.2 & 60.3 & 68.5 & 0.143 & 50.5 \\
\midrule
\hphantom{+}+ Backbone Fixed     & 60.5 & 87.6 & 68.8 & 73.0 & 0.135 & 53.1 \\
\hphantom{+}+ PEFT Fixed         & 59.2 & 86.5 & 62.0 & 68.7 & 0.143 & 49.4 \\
\hphantom{+}+ Probe Fixed        & 60.3 & 86.1 & 61.2 & 68.1 & 0.151 & 43.8 \\
\midrule
\textbf{VJEPA2 ViT-L}\\
Reference ($T{=}16$)            & 58.1 & 84.0 & 66.3 & 75.2 & 0.139 & 46.9 \\
Input-Fixed ($T_c{=}2$)         & 56.1 & 83.2 & 58.4 & 72.8 & 0.142 & 44.5 \\
\midrule
\hphantom{+}+ Backbone Fixed     & 58.3 & 83.2 & 65.3 & 74.0 & 0.123 & 46.6 \\
\hphantom{+}+ PEFT Fixed         & 57.1 & 83.8 & 61.5 & 71.3 & 0.143 & 44.7 \\
\hphantom{+}+ Probe Fixed        & 57.0 & 81.7 & 58.7 & 73.5 & 0.150 & 41.3 \\
\bottomrule
\end{tabular}
\end{adjustbox}
\end{table}

Across both experiments VJEPA2 reproduces the trends observed with InternVideo-Next. From Table~\ref{tab:rebuttal_vjepa2_fig3} we see that appearance and spatial tasks benefit primarily from more input frames, while motion tasks depend on temporal modeling. Table~\ref{tab:rebuttal_vjepa2_fig4} shows that temporal context should be allocated selectively to the backbone if needed. The main difference we observe between InternVideo-Next and VJEPA2 is on VSPW, where VJEPA2 shows a non-negligible temporal-modeling drop (5.2\%), while for InternVideo-Next the spatial degradation came entirely from reduced input frames. This suggests VJEPA2 relies somewhat more on backbone temporal modeling for dense prediction. However, the resulting recommendation to allocate temporal context to the backbone and not to the probe is unchanged.

\noindent \textbf{Image-pretrained backbone.} To test whether the temporal modeling in the probe harms spatial tasks with an image-pretrained backbone, we repeat the temporal context distribution experiment using DINOv3 ViT-B. Table~\ref{tab:rebuttal_spatial_comparison} reports the results for DINOv3 ViT-B, alongside InternVideo-Next and VJEPA2 on the spatial tasks. As in Section~\ref{sec:exp_context}, \textit{input-fixed} processes 16 input frames as independent two-frame segments in all components, and each ``+X Fixed'' variant keeps one component at the full 16-frame context while the other two remain at two frames. As DINOv3 encodes each frame independently ($T_{\mathrm{backbone}}{=}1$ always), we cannot run the ``+Backbone Fixed'' variant.

\begin{table}[h]
\centering
\caption{\textbf{How to Distribute Temporal Context, Spatial Tasks, Across Backbones.} InternVideo-Next ViT-B, VJEPA2 ViT-L, and DINOv3 ViT-B, each with ST-Adapter + Attentive Probe. Reference is $T{=}T_{\text{backbone}}{=}T_{\text{peft}}{=}T_{\text{probe}}{=}16$; Input-Fixed varies backbone/PEFT/probe context jointly at $T_c{=}2$; each ``+X Fixed'' variant keeps one component at $T{=}16$ while the other two shrink to $T_c{=}2$. DINOv3 has $T_{\text{backbone}}{=}1$ always (each frame encoded independently), so no ``+Backbone Fixed'' variant exists for it.}
\label{tab:rebuttal_spatial_comparison}
\renewcommand{\arraystretch}{1.15}
\setlength{\tabcolsep}{5pt}
\begin{adjustbox}{max width=\textwidth}
\scriptsize
\begin{tabular}{
  @{}l
  cc
  cc
  cc@{}
}
\toprule
& \multicolumn{2}{c}{\cellcolor{myblue}\textbf{InternVideo-Next ViT-B}}
& \multicolumn{2}{c}{\cellcolor{myblue}\textbf{VJEPA2 ViT-L}}
& \multicolumn{2}{c}{\cellcolor{myblue}\textbf{DINOv3 ViT-B}} \\
\cmidrule(lr){2-3}\cmidrule(lr){4-5}\cmidrule(lr){6-7}
& \makecell[c]{\textbf{ScanNet}\\\textit{\scriptsize AbsRel $\downarrow$}}
& \makecell[c]{\textbf{VSPW}\\\textit{\scriptsize mIoU $\uparrow$}}
& \makecell[c]{\textbf{ScanNet}\\\textit{\scriptsize AbsRel $\downarrow$}}
& \makecell[c]{\textbf{VSPW}\\\textit{\scriptsize mIoU $\uparrow$}}
& \makecell[c]{\textbf{ScanNet}\\\textit{\scriptsize AbsRel $\downarrow$}}
& \makecell[c]{\textbf{VSPW}\\\textit{\scriptsize mIoU $\uparrow$}} \\
\midrule
Reference ($T{=}16$)     & 0.140 & 49.5 & 0.139 & 46.9 & 0.135 & 54.6 \\
Input-Fixed ($T_c{=}2$)  & 0.143 & 50.5 & 0.142 & 44.5 & 0.133 & 55.0 \\
\midrule
\hphantom{+}+ Backbone Fixed  & 0.135 & 53.1 & 0.123 & 46.6 & \multicolumn{2}{c}{--} \\
\hphantom{+}+ PEFT Fixed      & 0.143 & 49.4 & 0.143 & 44.7 & 0.134 & 54.8 \\
\hphantom{+}+ Probe Fixed     & 0.151 & 43.8 & 0.150 & 41.3 & 0.145 & 54.6 \\
\bottomrule
\end{tabular}
\end{adjustbox}
\end{table}

The degradation from temporal aggregation in the probe indeed also occurs for the image-pretrained DINOv3: +Probe Fixed is the worst configuration on ScanNet (0.145 vs. 0.133 for input-fixed) and provides no benefit on VSPW. Across all three backbones, allocating long temporal context to the probe is therefore consistently unhelpful for spatial prediction.

\subsection{Inference Memory} \label{app:inferece_mem}
We measure peak inference memory in Table~\ref{tab:peft_efficiency}. It varies only modestly across PEFT methods and remains nearly constant across temporal context configurations, including long PEFT context. This is because input-fixed configurations process the same number of tokens regardless of where temporal context is allocated. Memory decreases only when the input itself is shortened in the matched setting, at a substantial cost to accuracy.

\begin{table}[h]
\centering
\caption{\textbf{Inference Cost of PEFT Methods.} Peak inference memory varies only modestly across PEFT paradigms. \textbf{Best} and \underline{second-best} are highlighted. \textsuperscript{\dag}Weights can be merged to obtain attentive probe throughput and memory. }
\vspace{-0.5em}
\label{tab:peft_efficiency}
\renewcommand{\arraystretch}{1.15}
\setlength{\tabcolsep}{5pt}

\begin{adjustbox}{max width=\textwidth}
\scriptsize
\begin{tabular}{
  @{}lll l
  cc
  cc
  c
  cc
  c@{}
}
\toprule
& & & & \multicolumn{2}{c}{\cellcolor{myblue}\textbf{Appearance}}
& \multicolumn{2}{c}{\cellcolor{myblue}\textbf{Motion}}
& &
\multicolumn{2}{c}{\cellcolor{myblue}\textbf{Spatial}} & \\
\cmidrule(lr){5-6}\cmidrule(lr){7-8}\cmidrule(lr){10-11}
& &
\makecell[l]{\textbf{Params}\\\textit{\scriptsize (M) $\downarrow$}}
& \makecell[l]{\textbf{Peak Mem}\\\textit{\scriptsize (GB) $\downarrow$}}
& \makecell[c]{\textbf{CAER}\\\textit{\scriptsize top-1 $\uparrow$}}
& \makecell[c]{\textbf{NurViD}\\\textit{\scriptsize top-1 $\uparrow$}}
& \makecell[c]{\textbf{IndustReal}\\\textit{\scriptsize top-1 $\uparrow$}}
& \makecell[c]{\textbf{MammAlps}\\\textit{\scriptsize top-1 $\uparrow$}}
& \makecell[c]{\textbf{Clips/s}\\\textit{\scriptsize $\uparrow$}}
& \makecell[c]{\textbf{ScanNet}\\\textit{\scriptsize AbsRel $\downarrow$}}
& \makecell[c]{\textbf{VSPW}\\\textit{\scriptsize mIoU $\uparrow$}}
& \makecell[c]{\textbf{Frames/s}\\\textit{\scriptsize $\uparrow$}} \\
\midrule
 \multirow{2}{*}{\rotatebox[origin=c]{90}{\textbf{Base}}} &\g Full Finetuning & \g 87.1 & \g \textbf{3.98} & \g 60.3\sd{0.6} & \g \textbf{86.5}\sd{0.7} & \g \underline{67.2}\sd{0.5} & \g \textbf{71.7}\sd{0.7} & \g \underline{265.5} & \g 0.167\sd{0.006} & \g 40.2\sd{0.7} & \g \textbf{2041.3} \\
  &Attentive Probe & \textbf{7.4} & 4.14 & 59.7\sd{0.8} & 85.6\sd{0.3} & 62.9\sd{0.9} & 69.4\sd{0.8} & 264.9 & 0.160\sd{0.004} & 40.9\sd{0.2} & \underline{1860.5} \\
\midrule
 \multirow{3}{*}{\rotatebox[origin=c]{90}{\textbf{Selective}}} &\g BitFit & \g 7.4 & \g \underline{4.14} & \g 60.4\sd{0.7} & \g 85.7\sd{1.0} & \g 63.7\sd{0.9} & \g 69.9\sd{1.3} & \g \textbf{265.7} & \g 0.153\sd{0.002} & \g 44.2\sd{0.5} & \g 1855.5 \\
  &LayerNorm & \underline{7.4} & 4.17 & 59.2\sd{1.0} & 85.7\sd{0.5} & 64.3\sd{0.3} & 69.3\sd{0.8} & 264.5 & 0.152\sd{0.003} & 44.4\sd{0.4} & 1859.2 \\
  &\g SSF & \g 7.6 & \g 4.92\textsuperscript{\dag} & \g 60.2\sd{1.2} & \g 85.7\sd{0.7} & \g 65.4\sd{0.6} & \g 67.4\sd{2.0} & \g 217.4\textsuperscript{\dag} & \g \textbf{0.146}\sd{0.003} & \g 47.0\sd{0.2} & \g 1555.1\textsuperscript{\dag} \\
\midrule
 \multirow{3}{*}{\rotatebox[origin=c]{90}{\textbf{Additive}}} &VPT & 8.4 & 4.26 & 59.7\sd{1.1} & 85.7\sd{0.4} & 65.3\sd{0.3} & 70.0\sd{1.0} & 243.9 & 0.154\sd{0.004} & 47.8\sd{0.4} & 1649.6 \\
  &\g VFPT & \g 8.8 & \g 4.34 & \g 59.4\sd{0.7} & \g 85.8\sd{0.6} & \g 64.4\sd{0.3} & \g 67.2\sd{0.8} & \g 233.0 & \g 0.161\sd{0.005} & \g 47.5\sd{1.2} & \g 1640.2 \\
  &BPT & 8.5 & 4.25 & 60.2\sd{0.5} & 85.7\sd{0.6} & 65.0\sd{1.0} & 70.0\sd{1.4} & 244.3 & 0.156\sd{0.005} & 48.7\sd{0.4} & 1549.3 \\
\midrule
 \multirow{2}{*}{\rotatebox[origin=c]{90}{\textbf{Adapt}}} &\g AdaptFormer & \g 11.3 & \g 4.20 & \g 60.2\sd{1.3} & \g 86.0\sd{0.6} & \g 66.7\sd{0.3} & \g \underline{71.1}\sd{0.5} & \g 255.9 & \g 0.151\sd{0.004} & \g \textbf{49.9}\sd{0.5} & \g 1808.8 \\
  &ST-Adapter & 14.1 & 4.59 & \textbf{60.9}\sd{0.8} & \underline{86.2}\sd{0.8} & 67.0\sd{0.3} & 69.6\sd{0.9} & 231.7 & \underline{0.148}\sd{0.003} & \underline{49.2}\sd{0.5} & 1692.9 \\
\midrule
 \multirow{2}{*}{\rotatebox[origin=c]{90}{\textbf{Rank}}} &\g LoRA & \g 11.7 & \g 4.16\textsuperscript{\dag} & \g \underline{60.7}\sd{1.0} & \g 86.0\sd{0.6} & \g \textbf{68.1}\sd{0.2} & \g 70.0\sd{2.5} & \g 244.4\textsuperscript{\dag} & \g 0.163\sd{0.010} & \g 47.3\sd{0.6} & \g 1728.7\textsuperscript{\dag} \\
  &DoRA & 13.3 & 4.58\textsuperscript{\dag} & 59.6\sd{0.4} & 85.2\sd{0.4} & 66.2\sd{0.3} & 66.6\sd{1.1} & 237.4\textsuperscript{\dag} & 0.169\sd{0.007} & 45.6\sd{0.5} & 1683.0\textsuperscript{\dag} \\
\bottomrule
\end{tabular}
\end{adjustbox}
\end{table}

\section{Per-dataset Graphs}
\label{app:per_dataset}

\begin{figure}
    \centering
    \includegraphics[width=\textwidth]{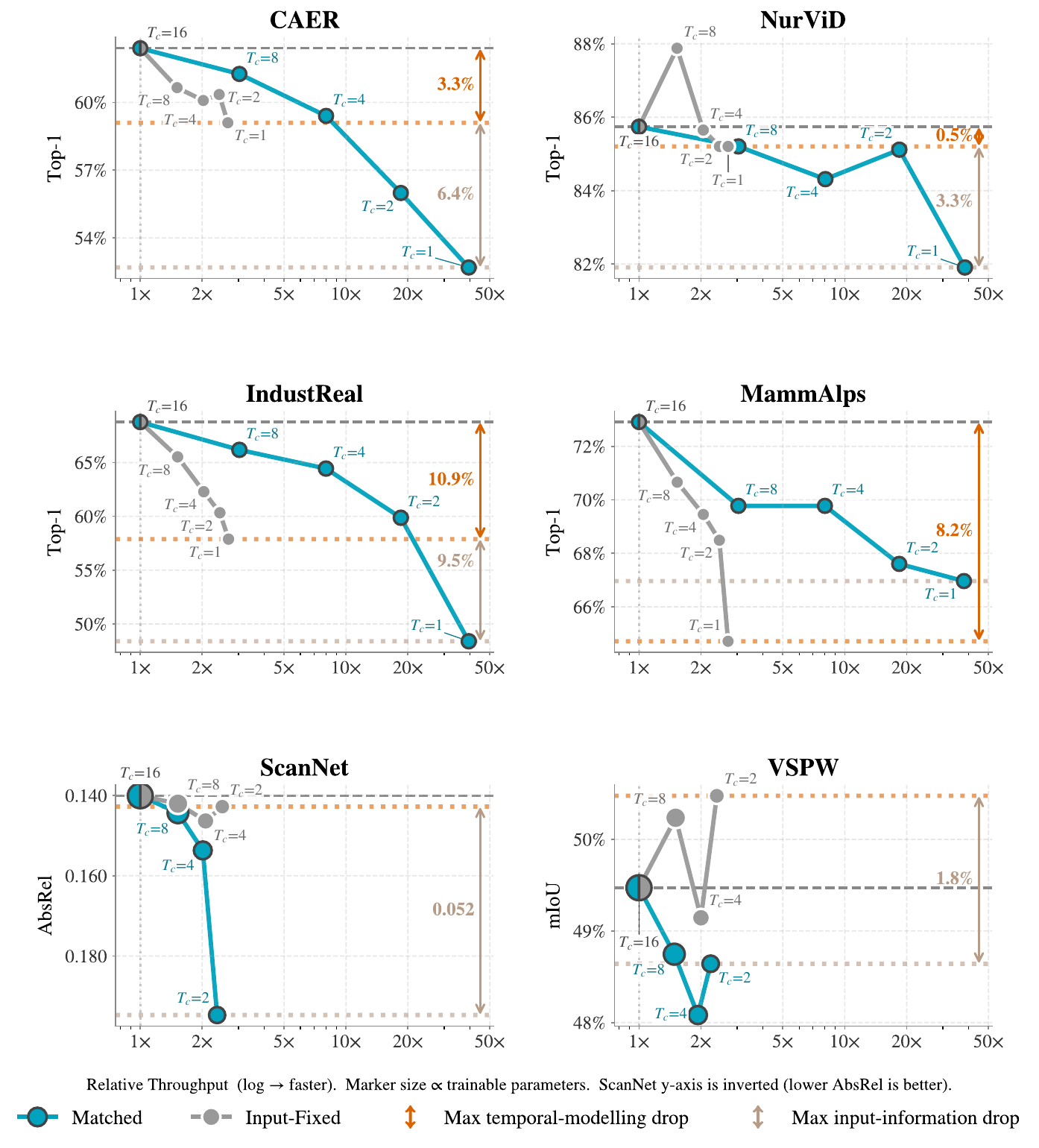}
    \caption{\textbf{Disentangling Temporal Modeling from Input Frames.} Decoupling the input frame count from the internal temporal context of the backbone, PEFT, and probe separates performance drops into a \emph{temporal-modeling} component and an \emph{input-information} component. Drops are measured at the smallest $T_c$ (1 for classification, 2 for spatial tasks) in the metric's own units. This is the per-dataset version of \autoref{fig:uniform_input}.}
    \label{fig:uniform_input_appendix}
\end{figure}

\begin{figure}
    \centering
    \includegraphics[width=\textwidth]{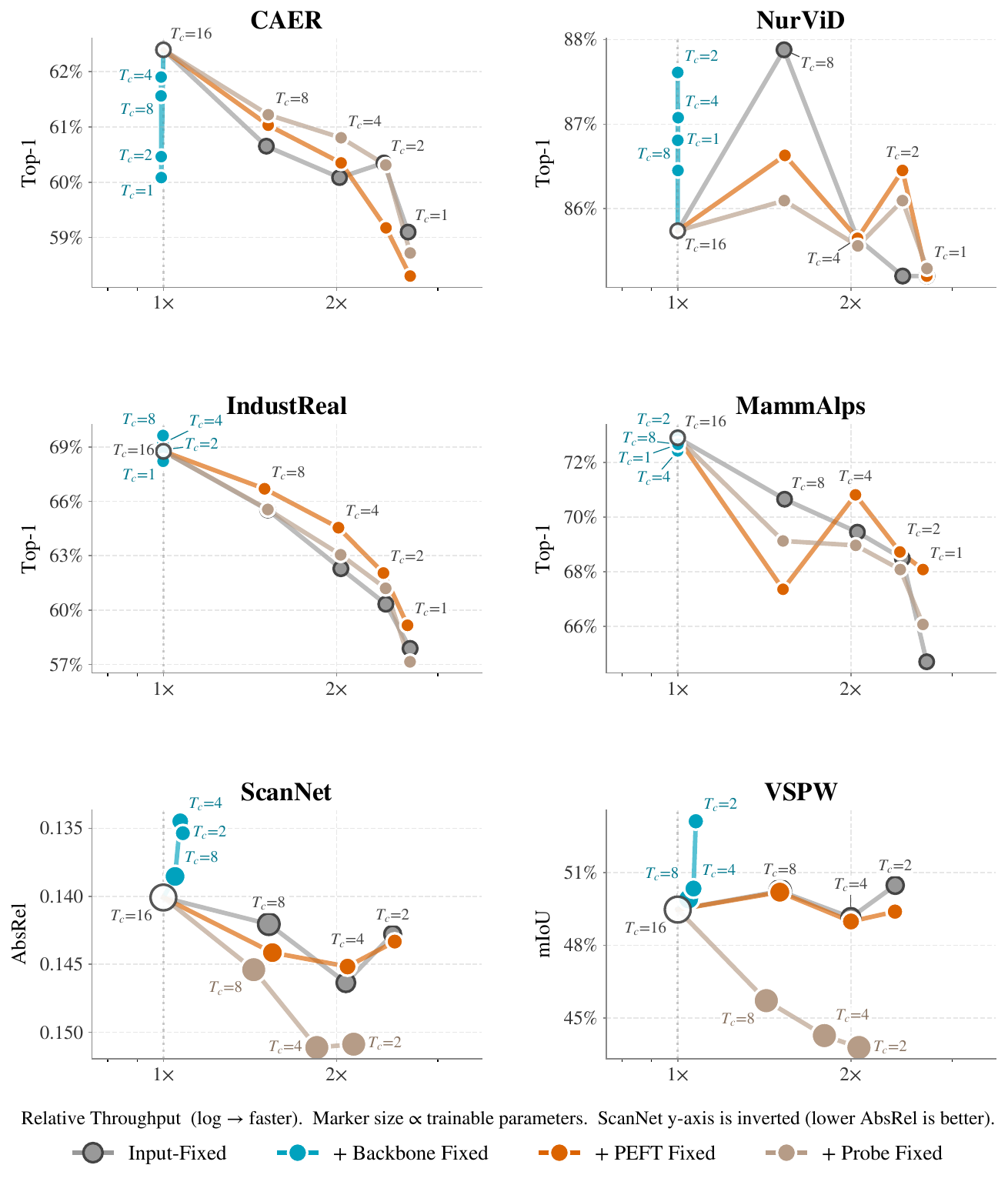}
    \caption{\textbf{How to Distribute Temporal Context.} Input-fixed varies the temporal context $T_c$ of backbone, PEFT, and probe jointly, while each ``+X Fixed'' curve keeps component X at 16 frames and varies the other two with $T_c$. A curve above input-fixed thus means that X benefits from temporal context. This is the per-dataset version of \autoref{fig:comp_modules}.}
    \label{fig:comp_modules_appendix}
\end{figure}

We provide per-dataset versions of the temporal modeling analyses presented in the main paper, where results were aggregated across the appearance, motion, and spatial categories. Figure~\ref{fig:uniform_input_appendix} reports the per-dataset breakdown of Figure~\ref{fig:uniform_input}. Figure~\ref{fig:comp_modules_appendix} provides the per-dataset counterpart of Figure~\ref{fig:comp_modules}. 

\section{Extended Dataset Selection Criteria}
\label{app:ext_data_sel}
Beyond the stated criteria in Section~\ref{sec:selection_criteria}, we selected two complementary datasets for each task type (appearance, motion and spatial), while ensuring that all datasets supported the same RGB input, probe comparisons and temporal context experiments. We initially considered UCF101~\citep{soomro2012ucf101}, HMDB51~\citep{kuehne2011hmdb}, and Diving48~\citep{li2018resound} as these are the standard benchmarks used to evaluate PEFT methods for video understanding. These were excluded as they are either relatively large or drawn from web-video distributions likely to resemble, and potentially overlap with, pretraining data. We also considered CholecTrack20~\citep{nwoye2025cholectrack20}, a low-resource surgical tracking dataset. However, this task requires initial object coordinates to be incorporated into the probe, preventing the evaluation of multiple probe designs as well as the temporal context experiments. For the same reason we excluded Waymo Open~\citep{sun2020scalability} and video object segmentation tasks. 

\section{Extended Broader Impact}
\label{app:broader}
Efficient adaptation of video foundation models can make advanced video understanding systems more accessible by reducing the computation and annotation costs required for deployment. This is particularly relevant for applications where labeled video data is scarce or expensive, such as healthcare, scientific imaging, manufacturing, robotics and environmental monitoring. By studying parameter-efficient adaptation in low-resource settings our work may help enable video models in settings where large-scale training or finetuning is impractical.

At the same time, improving the efficiency of video adaptation may also lower the barrier to deploying large-scale video analysis systems in privacy-sensitive settings, including surveillance and automated monitoring. As with other works based on foundation models, biases inherited from pretraining data may also transfer to downstream applications, particularly in low-resource domains where evaluation data is limited. Careful consideration of privacy, fairness and deployment constraints therefore remain important when applying our findings in practice.

\newpage
\section*{NeurIPS Paper Checklist}

\begin{enumerate}

\item {\bf Claims}
    \item[] Question: Do the main claims made in the abstract and introduction accurately reflect the paper's contributions and scope?
    \item[] Answer: \answerYes{} %
    \item[] Justification: The paper is an empirical study and the claims are justified by the results in Section~\ref{sec:experiments}.

\item {\bf Limitations}
    \item[] Question: Does the paper discuss the limitations of the work performed by the authors?
    \item[] Answer: \answerYes{} %
    \item[] Justification: The limitations are discussed in Section~\ref{sec:conclusion}.

\item {\bf Theory assumptions and proofs}
    \item[] Question: For each theoretical result, does the paper provide the full set of assumptions and a complete (and correct) proof?
    \item[] Answer: \answerNA{} %
    \item[] Justification: This is an empirical study, no theoretical results.

    \item {\bf Experimental result reproducibility}
    \item[] Question: Does the paper fully disclose all the information needed to reproduce the main experimental results of the paper to the extent that it affects the main claims and/or conclusions of the paper (regardless of whether the code and data are provided or not)?
    \item[] Answer: \answerYes{} %
    \item[] Justification: The experimental details are outlined in Appendix~\ref{app:experiment_details}.

\item {\bf Open access to data and code}
    \item[] Question: Does the paper provide open access to the data and code, with sufficient instructions to faithfully reproduce the main experimental results, as described in supplemental material?
    \item[] Answer: \answerYes{} %
    \item[] Justification: The link to the code can be found in the abstract, and the benchmark consists of existing research datasets.

\item {\bf Experimental setting/details}
    \item[] Question: Does the paper specify all the training and test details (e.g., data splits, hyperparameters, how they were chosen, type of optimizer) necessary to understand the results?
    \item[] Answer: \answerYes{} %
    \item[] Justification: They are outlined in Table~\ref{tab:benchmark} and Appendix~\ref{app:experiment_details}.

\item {\bf Experiment statistical significance}
    \item[] Question: Does the paper report error bars suitably and correctly defined or other appropriate information about the statistical significance of the experiments?
    \item[] Answer: \answerYes{} %
    \item[] Justification: Our main results in Table~\ref{tab:peft_results} contain error bars and the significance is discussed in the text. Error bars are sample standard deviations over 5 random seeds. Importantly, the PEFT comparison is the part of our study where many of the relevant performance differences are smallest. The observed seed variability is substantially smaller than the effects supporting the paper's other main conclusions. In particular, the temporal context experiments show large changes, while the principal backbone and probe trends recur across multiple datasets or configurations.

\item {\bf Experiments compute resources}
    \item[] Question: For each experiment, does the paper provide sufficient information on the computer resources (type of compute workers, memory, time of execution) needed to reproduce the experiments?
    \item[] Answer: \answerYes{} %
    \item[] Justification: As stated in Appendix~\ref{app:experiment_details} all experiments are done on 2$\times$H100 GPUs on a shared research cluster.
    
\item {\bf Code of ethics}
    \item[] Question: Does the research conducted in the paper conform, in every respect, with the NeurIPS Code of Ethics \url{https://neurips.cc/public/EthicsGuidelines}?
    \item[] Answer: \answerYes{} %
    \item[] Justification: Yes, we followed the code of ethics.

\item {\bf Broader impacts}
    \item[] Question: Does the paper discuss both potential positive societal impacts and negative societal impacts of the work performed?
    \item[] Answer: \answerYes{} %
    \item[] Justification: Broader impact is discussed in Section~\ref{sec:conclusion} and Appendix~\ref{app:broader}. %
    
\item {\bf Safeguards}
    \item[] Question: Does the paper describe safeguards that have been put in place for responsible release of data or models that have a high risk for misuse (e.g., pre-trained language models, image generators, or scraped datasets)?
    \item[] Answer: \answerNA{} %
    \item[] Justification: We do not pretrain large-scale models. Our adapters are inherently task-specific, and because the tasks we consider pose no safety risks, we do not implement additional safeguards.

\item {\bf Licenses for existing assets}
    \item[] Question: Are the creators or original owners of assets (e.g., code, data, models), used in the paper, properly credited and are the license and terms of use explicitly mentioned and properly respected?
    \item[] Answer: \answerYes{} %
    \item[] Justification: All models and datasets used in the paper are cited, and their licenses permit use for research purposes.

\item {\bf New assets}
    \item[] Question: Are new assets introduced in the paper well documented and is the documentation provided alongside the assets?
    \item[] Answer: \answerYes{} %
    \item[] Justification: We release no new data. We release our code under an MIT license with documentation and reproduction scripts.

\item {\bf Crowdsourcing and research with human subjects}
    \item[] Question: For crowdsourcing experiments and research with human subjects, does the paper include the full text of instructions given to participants and screenshots, if applicable, as well as details about compensation (if any)? 
    \item[] Answer: \answerNA{} %
    \item[] Justification: No crowdsourcing or experiments with human subjects were conducted.

\item {\bf Institutional review board (IRB) approvals or equivalent for research with human subjects}
    \item[] Question: Does the paper describe potential risks incurred by study participants, whether such risks were disclosed to the subjects, and whether Institutional Review Board (IRB) approvals (or an equivalent approval/review based on the requirements of your country or institution) were obtained?
    \item[] Answer: \answerNA{} %
    \item[] Justification: No experiments with human subjects were conducted.

\item {\bf Declaration of LLM usage}
    \item[] Question: Does the paper describe the usage of LLMs if it is an important, original, or non-standard component of the core methods in this research? Note that if the LLM is used only for writing, editing, or formatting purposes and does \emph{not} impact the core methodology, scientific rigor, or originality of the research, declaration is not required.
    \item[] Answer: \answerNA{} %
    \item[] Justification: LLMs were only used for grammatical editing.

\end{enumerate}

\end{document}